\documentclass[10pt]{article}

\usepackage[margin=0.78in]{geometry}
\usepackage{graphicx}
\usepackage{booktabs}
\usepackage{amsmath}
\usepackage{amssymb}
\usepackage{algorithm}
\usepackage{algpseudocode}
\usepackage{xcolor}
\usepackage{microtype}
\usepackage{caption}
\usepackage{hyperref}
\usepackage[numbers,sort&compress]{natbib}

\hypersetup{colorlinks=true,linkcolor=black,urlcolor=teal,citecolor=black}
\title{\textbf{Graph-structured joint-embedding prediction for single-cell transcriptomics}}
\author{Yuhao Wang$^{1,*}$,
Zelin Zang$^{1,2,*,\dagger}$,
Yuxuan Liu$^{1}$,
Zhen Lei$^{2,3,4,\dagger}$,
Stan Z. Li$^{1,\dagger}$\\[0.5em]
\small\parbox{0.96\textwidth}{\centering
$^{1}$Westlake University, Hangzhou, Zhejiang, China\\
$^{2}$Centre for Artificial Intelligence and Robotics (CAIR), Hong Kong Institute of Science \& Innovation, Chinese Academy of Sciences (HKISI-CAS), Hong Kong SAR, China\\
$^{3}$State Key Laboratory of Multimodal Artificial Intelligence Systems (MAIS), Institute of Automation, Chinese Academy of Sciences (CASIA), Beijing, China\\
$^{4}$School of Artificial Intelligence, University of Chinese Academy of Sciences (UCAS), Beijing, China\\[0.25em]
$^{*}$These authors contributed equally to this work.\\
$^{\dagger}$Correspondence: Zelin Zang
(\href{mailto:zangzelin@westlake.edu.cn}{zangzelin@westlake.edu.cn}), Zhen Lei
(\href{mailto:zhen.lei@ia.ac.cn}{zhen.lei@ia.ac.cn}) and Stan Z. Li
(\href{mailto:Stan.ZQ.Li@westlake.edu.cn}{Stan.ZQ.Li@westlake.edu.cn}).}}
\date{}

\newcommand{\method}{GB-JEPA}

\definecolor{BioMTeal}{HTML}{087F83}
\newcommand{\suppTableStyle}{\sffamily\footnotesize\renewcommand{\arraystretch}{1.10}\setlength{\tabcolsep}{4pt}}
\newcommand{\suppWideTableStyle}{\sffamily\scriptsize\renewcommand{\arraystretch}{1.08}\setlength{\tabcolsep}{3pt}}
\newcommand{\suppDenseTableStyle}{\sffamily\footnotesize\renewcommand{\arraystretch}{1.10}\setlength{\tabcolsep}{1.5pt}}
\newcommand{\biomrow}{}
\newcommand{\rankrow}{}

\begin{document}

\maketitle
\vspace{-1.5em}

\begin{abstract}
Single-cell transcriptomes are sparse, unordered measurements of coordinated biological activity, yet many self-supervised models define their learning target at the level of individual genes or reconstructed expression values. Here we introduce \method{}, a joint-embedding predictive architecture in which graph-connected gene blocks form the latent prediction units. A student encoder observes the genes outside each target block, whereas an exponential-moving-average teacher represents the full observed gene set. Teacher states are aggregated within a block before the prediction error is evaluated, converting thousands of sparse gene-level targets into a small collection of intermediate-scale latent targets. The graph therefore serves not only as prior information, but also as a mechanism for controlling the structure, scale and predictability of self-supervision. We evaluate the resulting frozen cell representations through objective diagnostics, cell-state annotation, continuous-expression and pathway probes, perturbation-response prediction, computational measurements and direct biological interrogation. This study presents a graph-structured route for adapting JEPA learning to molecular observations and a general framework for designing prediction units from relationships native to a biological modality.
\end{abstract}

Single-cell RNA sequencing observes each cell through a sparse and depth-dependent sample of its transcriptome. Cells in the same biological state can consequently contain different detected genes, while library size and detection depth can dominate apparent similarity\cite{lopez2018scvi}. The processes of interest are organized at a different scale. Cell identity, activation and response to perturbation arise from coordinated programmes in which multiple genes provide complementary and partly redundant evidence. Representation learning for single cells must therefore recover structured biological variation from incomplete molecular observations.

Pretrained transcriptomic models have shown that broad collections of cells can support reusable representations. Geneformer encodes ranked gene contexts, scGPT uses generative masked modelling, and scFoundation and scMulan combine value-aware representations with reconstruction or multitask objectives\cite{theodoris2023geneformer,cui2024scgpt,hao2024scfoundation,bian2024scmulan}. CellFM extends pretraining to more than 100 million cells, and Nicheformer integrates dissociated and spatial transcriptomes\cite{zeng2025cellfm,tejada2025nicheformer}. Their progress makes the choice of self-supervised target increasingly important. A model can reproduce individual measurements without necessarily organizing cells around the biological factors required by downstream analysis, and comparative studies report substantial variation across datasets and tasks\cite{kedzierska2025zeroshot,denadel2026scaling,ahlmann2025perturbation}.

Joint-embedding predictive architectures offer a different learning principle. Rather than reconstructing the input, a JEPA predicts the representation of a target from related context\cite{lecun2022jepa}. I-JEPA predicts latent representations of masked image regions, and V-JEPA extends this principle to spatiotemporal regions in video\cite{assran2023ijepa,bardes2024vjepa}. In these settings, a target is a coherent part of the underlying scene. The corresponding unit is not obvious for transcriptomics. Genes are unordered, individual observations are noisy and a single gene can participate in several biological processes. Treating each masked gene as the molecular analogue of an image patch therefore ignores the relational organization of the transcriptome.

\begin{figure}[t]
\centering
\includegraphics[width=\linewidth]{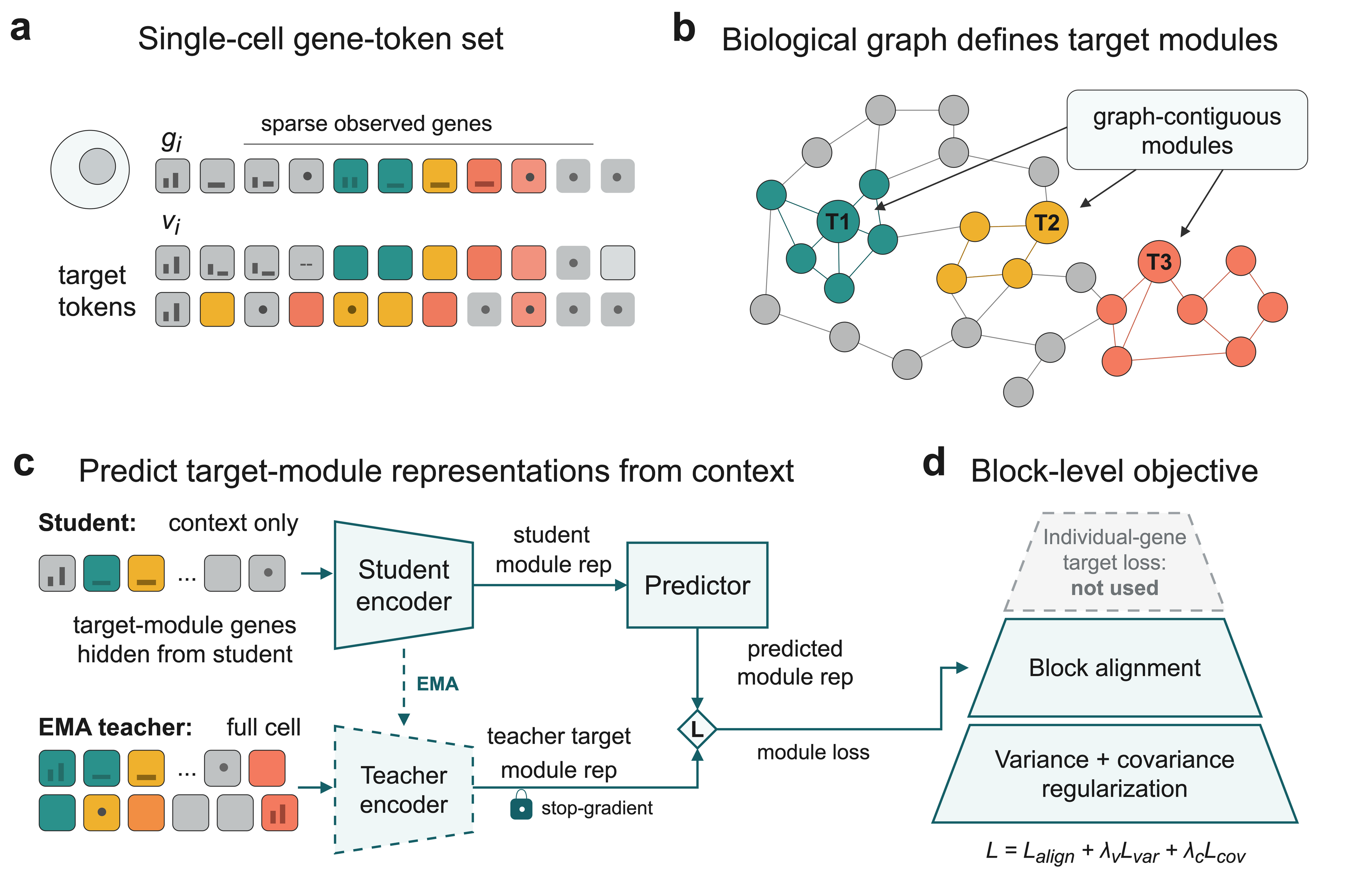}
\caption{\textbf{Graph-connected block prediction with \method{}.}
\textbf{a}, A cell is represented as a sparse set of observed genes and their expression values.
\textbf{b}, A gene graph defines connected candidate target blocks.
\textbf{c}, The student encodes the observed complement of each target block, whereas the EMA teacher represents the full observed gene set. The student predicts the aggregate teacher representation of each target block from this context, and the teacher target is stop-gradient.
\textbf{d}, \method{} aggregates target-gene states into one representation per block before evaluating a block-level latent prediction error. Detailed definitions and training pseudocode are provided in Supplementary Methods.}
\label{fig:method}
\end{figure}

We address this problem with \method{}, a gene-block joint-embedding predictive architecture for single-cell transcriptomics (Fig.~\ref{fig:method}). A sparse graph combines high-confidence STRING protein associations with coexpression estimated from the unlabeled pretraining corpus\cite{szklarczyk2023string}. Breadth-first expansion from an observed seed defines a graph-connected candidate block. The student encodes the complementary observed genes, and an exponential-moving-average teacher encodes the full observed gene set. Teacher states belonging to the block are aggregated before the prediction error is evaluated, so each sampled block contributes one latent target rather than a collection of independent gene targets. The graph thus defines both relational coordinates and a distribution over intermediate-scale prediction units.

We use \emph{graph-connected gene block} for the operational target sets sampled during pretraining. \emph{Biological programme} and \emph{pathway} denote gene sets supported by independent annotation or downstream analysis. This distinction motivates a direct study of target design: graph topology, block scale, neighbourhood coverage and latent predictability can each change what the student is asked to infer.

The student, teacher and predictor use linear attention, which contracts keys and values before applying each query and avoids materializing a dense gene-by-gene attention matrix\cite{katharopoulos2020linear}. We examine the method through matched objective and graph controls, frozen-representation benchmarks, continuous-expression and perturbation-response probes, computational measurements and direct biological interrogation. Together, these analyses test whether graph-connected blocks provide useful latent targets, which properties of the target graph matter, and what biological information remains accessible from the resulting cell representation.

\section{Results}

\subsection{Token prediction can optimize without producing a robust cell representation}

We first asked whether successful latent prediction at the gene level is sufficient to organize cells in a biologically useful representation space. We trained \method{}, pure token-IJEPA and a decoder-only control using the same vocabulary, hidden width, linear-attention backbone and pretraining data. Token-IJEPA predicted individual teacher gene states, the decoder reconstructed expression, and \method{} predicted one teacher representation for each graph-defined target block.

\begin{figure}[t]
\centering
\includegraphics[width=\linewidth]{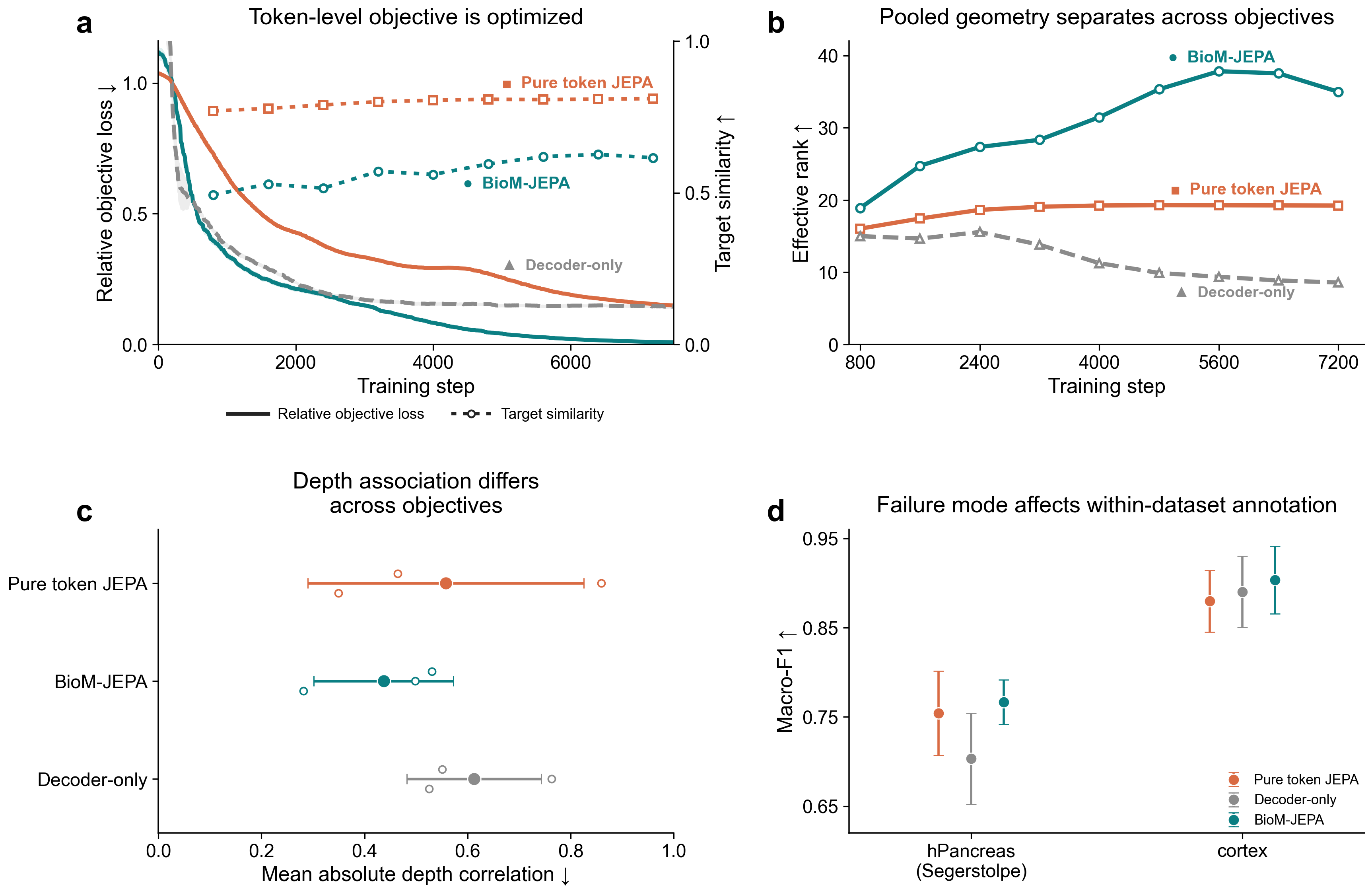}
\caption{\textbf{Objective optimization separates from pooled representation quality.}
\textbf{a}, Relative training losses for all three models (solid lines) and teacher-target similarity for the two JEPA models (dotted lines). Loss is normalized within each objective family.
\textbf{b}, Effective rank of hPancreas cell embeddings across training snapshots.
\textbf{c}, Mean absolute association between embedding diagnostics and detected-gene depth; open symbols show individual diagnostics and filled symbols their mean.
\textbf{d}, CellBench within-dataset Top-5 few-shot macro-$F_1$ for hPancreas and cortex (mean and s.d. over five seeds).}
\label{fig:failure}
\end{figure}

All three training losses decreased, while teacher-target similarity increased for the two JEPA models (Fig.~\ref{fig:failure}a). Their pooled cell representations nevertheless developed differently. Across matched hPancreas training snapshots, the effective rank of the \method{} embedding rose from approximately 19 to nearly 40. Token-IJEPA remained near 19, whereas the decoder-only representation contracted to fewer than ten effective dimensions (Fig.~\ref{fig:failure}b). Accurate token prediction or expression reconstruction therefore did not ensure that cell-to-cell variation was retained across many independent directions.

The geometry also differed in its association with sequencing depth. We correlated the leading embedding axis and the embedding norm with the number of detected genes. \method{} showed the weakest mean absolute association, token-IJEPA the strongest and decoder-only an intermediate value (Fig.~\ref{fig:failure}c). When the encoders were frozen and evaluated using the same CellBench Top-5 probe, \method{} achieved the highest macro-$F_1$ on both hPancreas and cortex (Fig.~\ref{fig:failure}d). These results support a failure mode specific to pooled transcriptomic representations: a gene-level predictive objective can converge while the resulting cell embedding remains low dimensional and coupled to technical depth.

\subsection{GB-JEPA improves the annotation--robustness balance}

\begin{figure}[t]
\centering
\includegraphics[width=\linewidth]{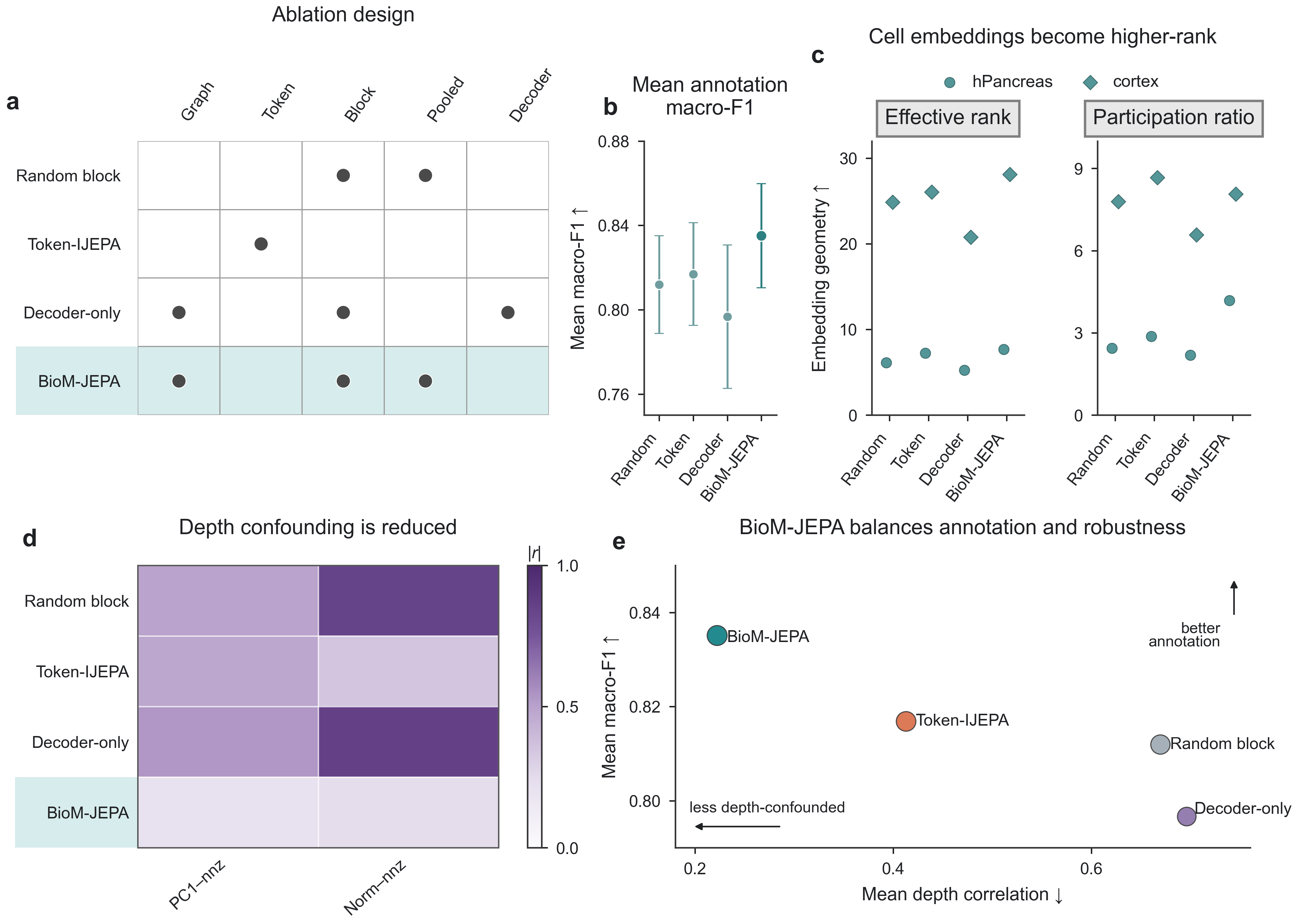}
\caption{\textbf{Controlled comparison of pretraining objective families.}
\textbf{a}, Components used by the four pretrained variants.
\textbf{b}, Mean Top-5 few-shot macro-$F_1$ over hPancreas and cortex.
\textbf{c}, Effective rank and participation ratio for hPancreas (circles) and cortex (diamonds).
\textbf{d}, Absolute correlations between detected-gene count and the leading embedding axis or embedding norm.
\textbf{e}, Mean within-dataset macro-$F_1$ versus mean depth association. Higher annotation performance and lower association are preferred.}
\label{fig:controls}
\end{figure}

We next compared graph-based target construction and objective choice using a shared encoder backbone and an identical embedding-extraction and downstream-evaluation pipeline. Random block sampled target sets uniformly from the vocabulary over the same size range. Token-IJEPA retained the teacher--student architecture and predicted genes individually. Decoder-only retained graph blocks and optimized expression reconstruction. \method{} combined graph-connected blocks with aggregate block-level prediction (Fig.~\ref{fig:controls}a).

\method{} achieved the highest mean Top-5 macro-$F_1$ across hPancreas and cortex (0.835), compared with 0.817 for token-IJEPA, 0.812 for random block and 0.797 for decoder-only (Fig.~\ref{fig:controls}b). The same ordering was observed for effective rank on both datasets (Fig.~\ref{fig:controls}c). Participation ratio, which measures how evenly variation is distributed across the representation, likewise placed \method{} above random block and decoder-only, with greater dataset dependence for token-IJEPA.

The complete block objective also reduced technical coupling in the tested diagnostics. Correlations between detected-gene count and either the leading embedding axis or embedding norm were lower for \method{} than for the three controls (Fig.~\ref{fig:controls}d). Plotting within-dataset annotation performance against depth association placed \method{} in the favourable region of high annotation score and low technical coupling (Fig.~\ref{fig:controls}e). Random targets produced weaker annotation despite matching the block-size range, while token prediction retained stronger depth association. Because all models used the same encoder backbone, the observed differences were associated with the complete pretraining formulations rather than encoder width or depth.

\subsection{Linear attention accelerates embedding extraction and fine-tuning}

The model contains twelve linear-attention encoder layers and a four-layer predictor. The attention core contracts keys and values into a shared summary before applying each query, so computation grows linearly with the number of observed genes at fixed model width\cite{katharopoulos2020linear}. This formulation processes the full observed-gene sequence for each cell.

We first isolated frozen encoder execution on identical ordered hPancreas cells at batch size 8. \method{} encoded 678.26 cells per second, compared with 353.35 cells per second for scFoundation, a 1.92-fold increase. hPancreas contained a median of 319 and a maximum of 1,659 observed gene tokens, all of which were retained during encoding.

We then measured a task-realistic one-epoch hPancreas run at the same batch size. The final encoder block, leaf normalization parameters and an identical two-layer prediction head were trainable, giving 26.09 million trainable parameters for each model. \method{} processed 24.37 training cells per second, compared with 4.24 for scFoundation, a 5.75-fold increase (Table~\ref{tab:efficiency}). In evaluation mode, held-out embedding throughput was 62.78 versus 16.68 cells per second, a 3.76-fold increase. The complete train--validation--test epoch required 66.08\,s for \method{} and 352.68\,s for scFoundation. Within this matched protocol, \method{} was faster than scFoundation in both frozen and partially tuned downstream use.

Strong frozen-probe performance emerged after 10.24 million cell presentations, approximately 46\% of the 22.1-million-cell local collection. The reported corpus sizes of Geneformer, scGPT, scFoundation and CellFM range from approximately 30 million to 102.3 million cells\cite{theodoris2023geneformer,cui2024scgpt,hao2024scfoundation,zeng2025cellfm}. GB-JEPA therefore learned a useful block-aware representation after a comparatively compact training exposure.

\begin{table}[t]
\centering
\caption{\textbf{Matched one-epoch hPancreas efficiency.} Both models used batch size 8, bfloat16 precision, one A100-SXM4-80GB GPU, the same data order, pooling structure and prediction head. Held-out throughput is measured during evaluation-mode validation and test loops.}
\label{tab:efficiency}
\begin{tabular}{lrrrrr}
\toprule
Model & Trainable & Train cells s$^{-1}$ & Held-out emb. cells s$^{-1}$ & Full epoch (s) & Peak (GiB) \\
\midrule
\method{} & 26,088,205 & 24.37 & 62.78 & 66.08 & 27.98 \\
scFoundation & 26,104,077 & 4.24 & 16.68 & 352.68 & 24.59 \\
\midrule
Speed-up & --- & \textbf{5.75$\times$} & \textbf{3.76$\times$} & \textbf{5.34$\times$} & --- \\
\bottomrule
\end{tabular}
\end{table}

\subsection{Frozen embeddings retain expression, pathway and neighbourhood information}

\begin{figure}[t]
\centering
\includegraphics[width=\linewidth]{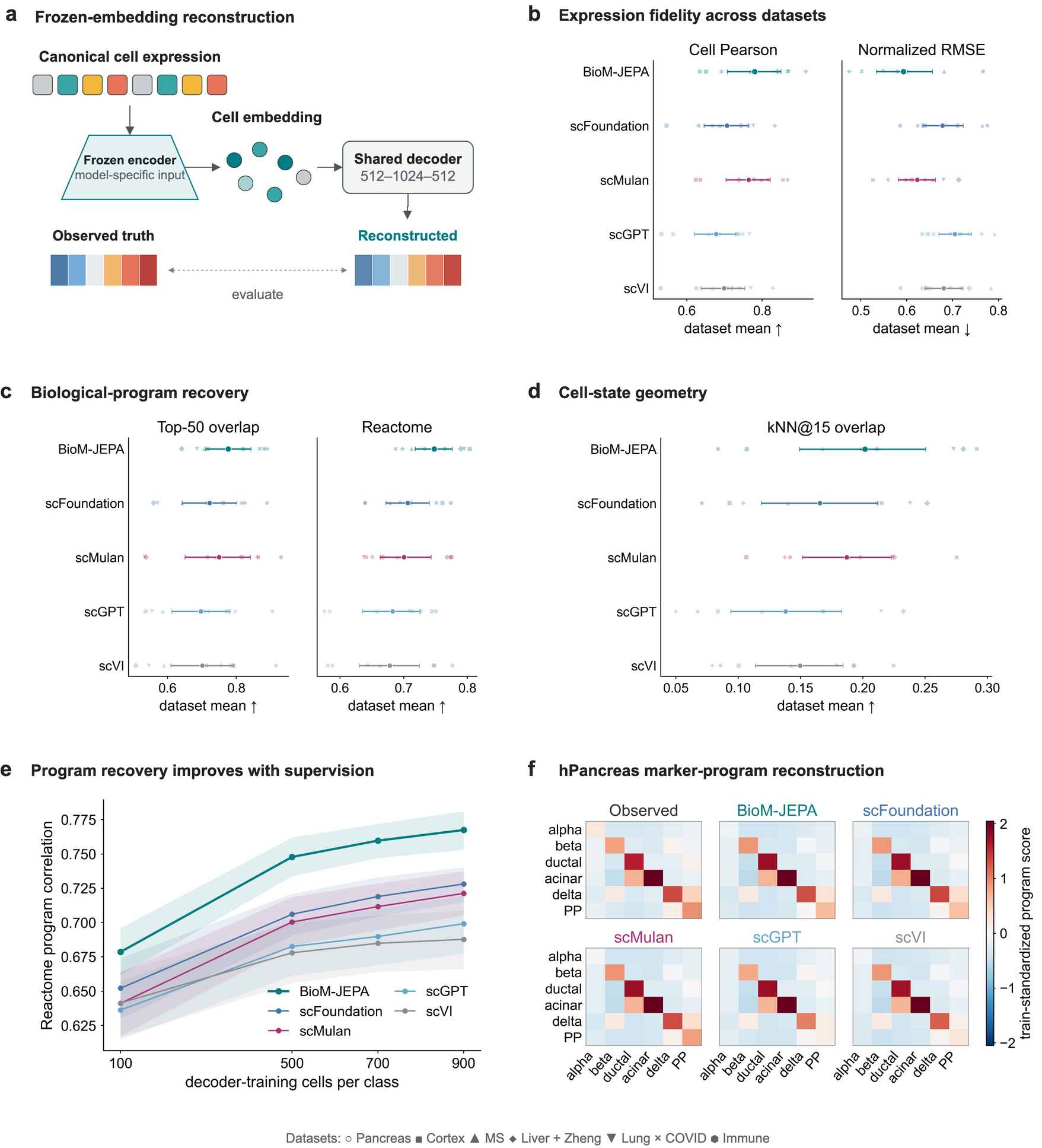}
\caption{\textbf{Reconstruction probes of frozen single-cell embeddings.}
\textbf{a}, Shared-decoder evaluation.
\textbf{b}, Cell-wise expression fidelity across eight datasets.
\textbf{c}, Recovery of highly expressed genes and Reactome programmes.
\textbf{d}, Preservation of cell-state neighbourhoods.
\textbf{e}, Reactome programme correlation as decoder supervision increases.
\textbf{f}, hPancreas marker-programme reconstruction. Small symbols in \textbf{b--d} denote datasets; large points and intervals summarize the dataset mean and its uncertainty.}
\label{fig:reconstruction}
\end{figure}

A representation can support cell-type classification while discarding continuous transcriptional structure. To test what remained accessible from each frozen embedding, we trained the same supervised decoder to predict a canonical expression vector (Fig.~\ref{fig:reconstruction}a). Encoder weights remained fixed, and every model used the same decoder architecture, supervision budget, held-out split and stopping rule.

Across eight CellBench datasets, \method{} achieved the highest dataset-mean cell-wise Pearson correlation and the lowest normalized root-mean-square error among the evaluated frozen representation models (Fig.~\ref{fig:reconstruction}b). It also recovered more of the 50 most highly expressed genes, gave the highest Reactome programme correlation and best preserved expression-space nearest neighbours (Fig.~\ref{fig:reconstruction}c,d). These metrics measure distinct properties: numerical expression fidelity, recovery of prominent genes, pathway-level activity and local cell-state geometry.

Increasing the decoder training budget improved programme recovery for every model, but \method{} remained strongest from 100 to 900 labelled cells per class (Fig.~\ref{fig:reconstruction}e). In hPancreas, reconstructed marker-programme matrices preserved the major endocrine and exocrine organization, including alpha, beta, ductal, acinar, delta and pancreatic-polypeptide states (Fig.~\ref{fig:reconstruction}f). Together, these results show that the frozen representation retained continuous and multiscale biological information alongside cell-identity information.

\subsection{Block representations support perturbation-response prediction}

\begin{figure}[t]
\centering
\includegraphics[width=\linewidth]{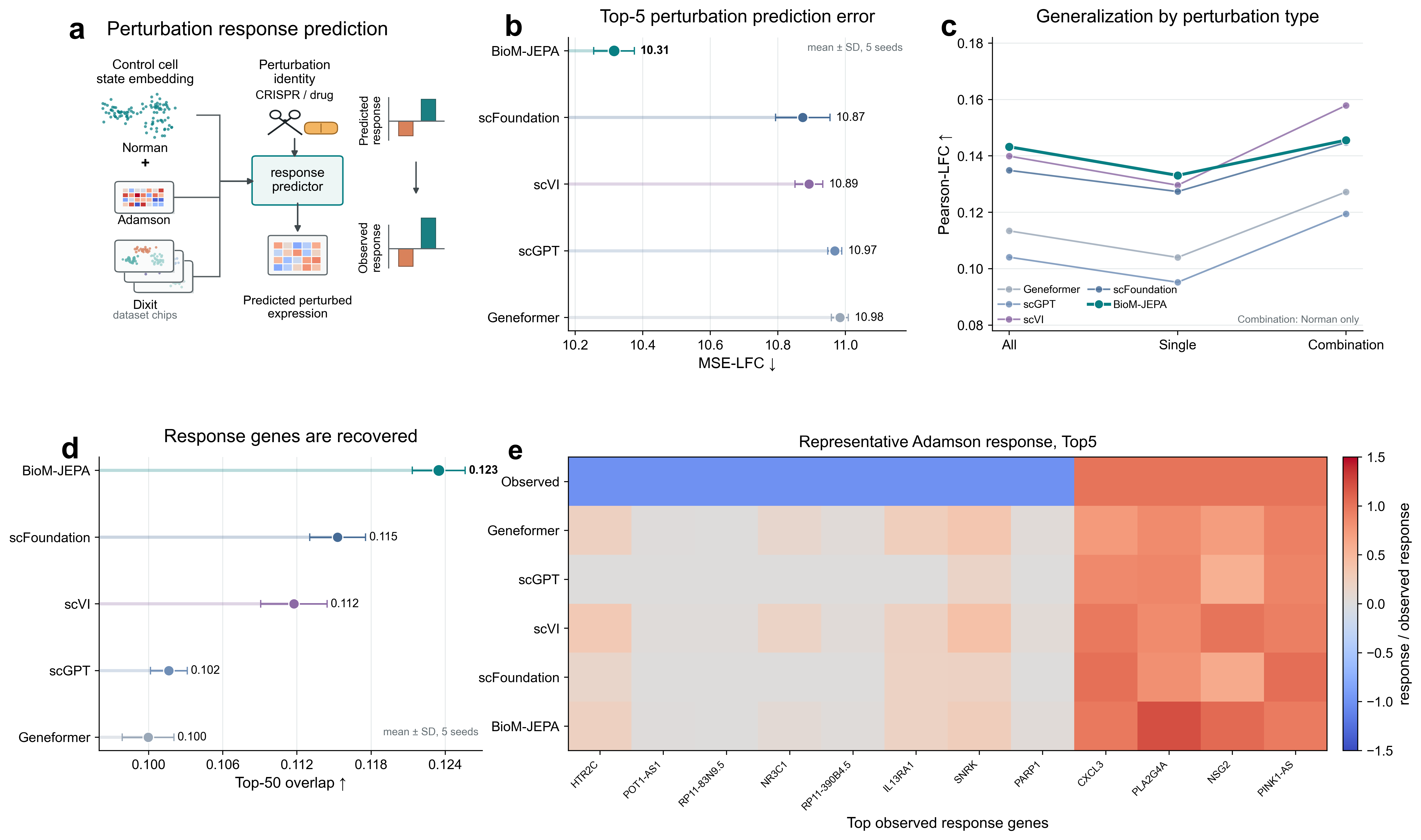}
\caption{\textbf{Perturbation prediction from frozen cell representations.}
\textbf{a}, CellBench response-prediction protocol.
\textbf{b}, Top-5 MSE-LFC (mean $\pm$ s.d. over five seeds; lower is better).
\textbf{c}, Pearson-LFC by perturbation type; combination perturbations are available for Norman only.
\textbf{d}, Top-50 response-gene overlap (mean $\pm$ s.d. over five seeds; higher is better).
\textbf{e}, Representative Adamson Top-5 observed and predicted response profiles.}
\label{fig:perturbation}
\end{figure}

We next evaluated perturbation-response prediction from frozen cell representations. The CellBench response model receives a control-cell embedding and a perturbation identity and predicts the perturbed expression profile (Fig.~\ref{fig:perturbation}a). We evaluated Adamson, Norman and Dixit using matched Top-$k$ controls and five fixed seeds. The encoder remained frozen and the response predictor was identical across models.

With Top-5 controls, \method{} achieved the lowest aggregate mean-squared error of the log-fold-change response (10.31), compared with 10.87--10.98 for scFoundation, scVI, scGPT and Geneformer (Fig.~\ref{fig:perturbation}b). Its Pearson correlation was highest for the aggregate and unseen-single categories (Fig.~\ref{fig:perturbation}c). On the Norman combination subset, the only dataset containing combination perturbations, scVI obtained the highest correlation.

\method{} also recovered the largest fraction of the top 50 response genes (0.123), followed by scFoundation (0.115; Fig.~\ref{fig:perturbation}d). In a representative Adamson condition, its predictions reproduced the direction and relative magnitude of the dominant response programme (Fig.~\ref{fig:perturbation}e). Agreement across error, correlation and response-gene recovery is consistent with the block representation preserving information relevant to both the magnitude and biological composition of perturbation responses in the tested protocol.

\subsection{GB-JEPA representations reflect biological programmes and genetic relationships}

\begin{figure}[t]
\centering
\includegraphics[width=\linewidth]{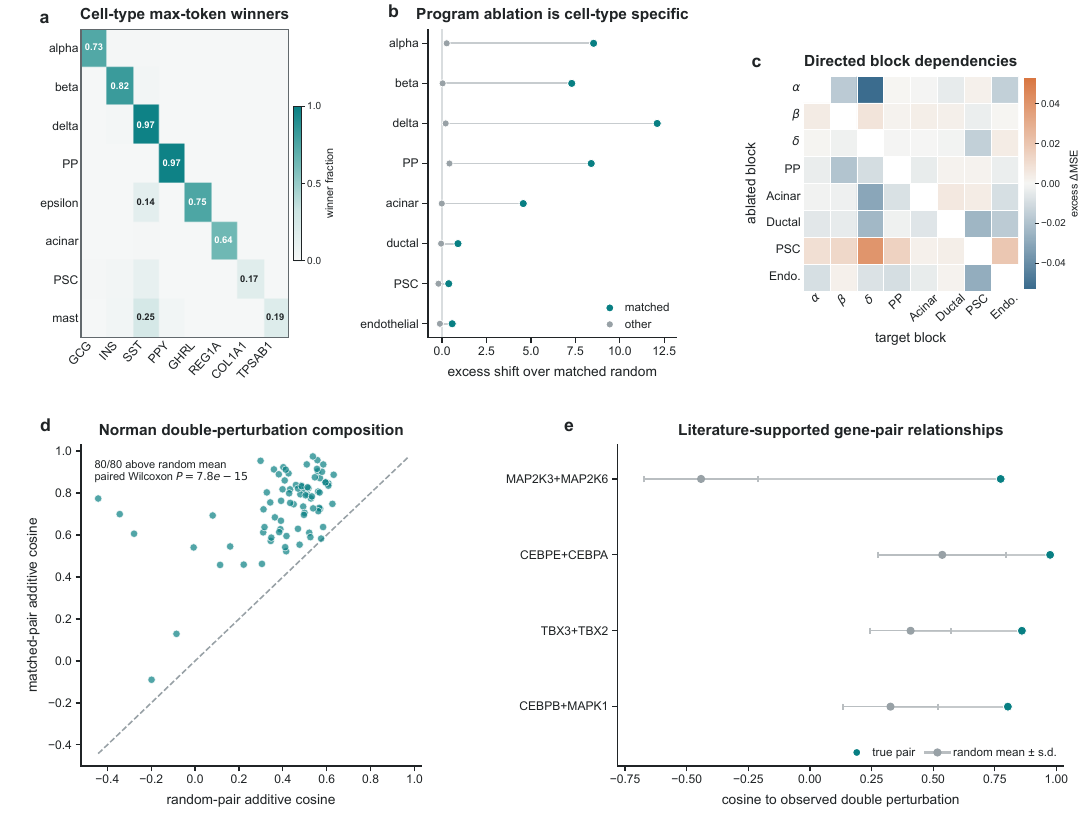}
\caption{\textbf{Biological structure reflected in frozen \method{} representations.}
\textbf{a}, Cell-type-resolved max-token winner fractions for canonical hPancreas genes.
\textbf{b}, Excess embedding shift after graph-programme ablation relative to one expression- and detection-matched random set per programme. Points summarize 24 cells per identity; the aggregate test uses the eight programme identities as paired statistical units.
\textbf{c}, Descriptive directed change in target-block prediction error after ablating each context programme, corrected by the mean of three matched-random ablations; each ordered pair was evaluated in 24 cells from the target-matched identity.
\textbf{d}, Agreement between each of 80 Norman double-perturbation displacements and the additive displacement of its matched constituent pair, compared with 200 matched-random pairs per condition.
\textbf{e}, Selected literature-supported gene pairs. Teal points show the matched pair; grey points and intervals show the random-pair mean and s.d. MAP2K3--MAP2K6, CEBPE--CEBPA, TBX3--TBX2 and CEBPB--MAPK1 represent p38 signalling, granulocytic differentiation, developmental regulation and ERK--C/EBP$\beta$ signalling, respectively. Complete definitions and source-data construction are provided in Supplementary Methods.}
\label{fig:biology}
\end{figure}

We finally interrogated the frozen representation directly for recognizable biological structure. In the hPancreas dataset\cite{segerstolpe2016pancreas}, the genes that most frequently supplied the maximal token activation aligned with canonical cell identities: \textit{GCG} in alpha cells, \textit{INS} in beta cells, \textit{SST} in delta cells, \textit{PPY} in pancreatic-polypeptide cells, \textit{GHRL} in epsilon cells and \textit{REG1A} in acinar cells (Fig.~\ref{fig:biology}a).

We then removed each of eight graph-defined programmes and measured how far the frozen cell embedding moved relative to removal of an expression- and detection-matched random gene set. Across the eight pre-specified programme identities, the corresponding cell identity showed a larger displacement than the mean of the other identities (mean excess shift, 5.34 versus 0.064; two-sided paired Wilcoxon test across programmes, $n=8$, $P=0.0078$; Fig.~\ref{fig:biology}b). Directed ablations also produced an asymmetric matrix of changes in target-block prediction error (Fig.~\ref{fig:biology}c). These results associate cell identity with coordinated graph programmes rather than only with isolated high-activation genes.

The Norman perturbation atlas provides a second test of biological organization\cite{norman2019genetic}. For each double perturbation, we compared its latent displacement with the sum of the displacements produced by its two constituent single perturbations. All 80 evaluated combination conditions agreed more strongly with their matched constituent pair than with the mean of 200 expression-matched random pairs per condition (two-sided paired Wilcoxon test across conditions, $n=80$, $P=7.85\times10^{-15}$; Fig.~\ref{fig:biology}d). The highest-agreement examples included MAP2K3--MAP2K6, which converge on p38 signalling\cite{raingeaud1996mkk}; CEBPA--CEBPE, which coordinate granulocytic differentiation\cite{avellino2022cebpa}; and the related developmental regulators TBX2--TBX3\cite{singh2012tbx}. These examples are consistent with established functional relationships being reflected in the geometry of the frozen representation and provide a starting point for testable biological hypotheses.

\clearpage

\section{Discussion}

This work introduces graph-connected gene blocks as prediction units for joint-embedding learning in single-cell transcriptomics. The central design choice is the order of operations: teacher states are first aggregated within a sampled block, and the student is then trained to predict that block representation from non-target genes. This changes the supervision from many sparse gene-level targets to a small number of structured latent targets. The resulting framework places target design alongside architecture and data scale as a primary determinant of a transcriptomic representation.

Intermediate-scale targets are well suited to sparse molecular observations. Failure to detect one gene provides limited evidence about the activity of the processes in which it participates, whereas a connected set can collect partially redundant evidence from several observed genes. Graph expansion also gives each target a reproducible identity without imposing an arbitrary linear order on the gene vocabulary. In this sense, graph-connected blocks play a role analogous to coherent spatial regions in I-JEPA and spatiotemporal regions in V-JEPA, while adapting the prediction unit to the organization of molecular data\cite{assran2023ijepa,bardes2024vjepa}.

The graph acts as a self-supervised curriculum: it determines which genes co-occur in a target, how rapidly a block expands, how much two targets overlap and how predictable a hidden block is from its remaining neighbourhood. Biological coherence and exploration efficiency can reinforce one another at some block scales and compete at others. This perspective makes graph construction an experimentally testable target-design problem.

Frozen-representation analyses connect this learning principle to cellular phenotypes. Annotation and clustering test whether cell identities are separated; reconstruction and pathway probes test whether continuous transcriptional information remains accessible; perturbation prediction tests whether the representation supports response modelling beyond cell-type labels. Direct interrogation adds a complementary layer: token-level activation, programme ablation and perturbation displacement can be used to identify relationships that are consistent with established pancreatic programmes and genetic interactions. These analyses are most informative when read together, because no single diagnostic is sufficient to establish biological usefulness.

Linear attention addresses the computational side of the same problem. It contracts key--value information before applying each query, while the graph-block objective specifies what information should be predicted. The measured throughput comparison characterizes the current \method{} and scFoundation implementations under a matched hPancreas protocol. Together, structured targets and efficient sequence processing make block-level JEPA practical at transcriptome scale.

The scope of the present evidence is defined by three features of the study design. Graph-connected blocks are operational prediction targets, whereas biological programmes are assigned through independent analyses. Corpus curation combined source-accession exclusion with exact cell-fingerprint screening of the available standardized parent matrices. Objective controls compare complete training formulations, and technical association is evaluated through the specified leading-axis and embedding-norm diagnostics.

\method{} extends joint-embedding prediction from ordered visual regions to unordered molecular observations. Its broader implication is that a JEPA target should follow the relational structure and semantic scale of the modality. Gene blocks instantiate this principle by linking molecular relations to the unit of self-supervision.

\begingroup
\small
\section{Methods}

\subsection{Problem formulation}

Let $\mathcal V$ denote a shared vocabulary of $G=19{,}264$ genes. A cell $i$ is represented by its observed gene set $\mathcal O_i\subseteq\mathcal V$ and normalized expression values $x_{ig}$ for $g\in\mathcal O_i$. \method{} learns from unlabeled cells by predicting latent representations of structured subsets of the observed transcriptome.

The central prediction unit is a \emph{graph-connected gene block}. We use this term strictly for a connected set sampled from a gene graph; biological programme or pathway refers only to a set supported by an independent annotation or downstream analysis. This distinction separates the mathematical object used for self-supervision from the biological interpretation tested after training.

\subsection{Cell representation and gene graph}

Raw counts were mapped to $\mathcal V$, normalized to a constant library size and transformed as $x_{ig}=\log(1+10^4c_{ig}/\sum_{g'}c_{ig'})$. Two technical values encoding a fixed resolution and log library depth were available to the encoders as context but were excluded from target sampling and downstream pooling.

We constructed a sparse gene graph $\mathcal G=(\mathcal V,\mathcal E)$ from two complementary sources. High-confidence STRING v12 protein associations supplied experimentally and computationally supported molecular links, and corpus-level coexpression supplied transcriptomic proximity\cite{szklarczyk2023string}. For each gene, we retained at most 64 outgoing neighbours after taking the binary union of STRING associations with combined score at least 700 and the strongest approximate absolute coexpression neighbours. The resulting adjacency contained 1,002,591 directed entries, with mean out-degree 52.04. Graph construction used these molecular associations and unlabeled corpus expression as its edge sources.

\subsection{Graph-connected target blocks}

For each cell, we request four target blocks. An observed seed gene and a block size between 512 and 2,000 genes are sampled for each target, and breadth-first expansion from the seed produces a candidate block $B_{ik}\subseteq\mathcal V$. Genes detected in the current cell define its realized target support, $T_{ik}=B_{ik}\cap\mathcal O_i$. The student receives the complementary observed set, $\mathcal C_i=\mathcal O_i\setminus\bigcup_kT_{ik}$, and the teacher receives the full observed gene set. Target construction is therefore shared across cells through the graph, whereas the realized target depends on the genes observed in each cell.

The breadth-first-search seed identifies the block. Its learned query combines the seed embedding with a mask embedding. On the student side, the query selects genes from the complementary observed set that fall within a 64-neighbour coexpression neighbourhood around the seed. Sampling reserves an observed neighbour in this seed-local context for every accepted block. A temperature-controlled weighting summarizes the neighbourhood into one student block state.

\subsection{Student--teacher prediction}

The student encoder $f_{\theta}$ and teacher encoder $f_{\xi}$ share a 12-layer, 768-dimensional linear-attention architecture. The teacher parameters follow the exponential-moving-average update $\xi\leftarrow\mu\xi+(1-\mu)\theta$, with momentum increasing from 0.996 to 0.9997. A four-layer predictor $p_{\phi}$ maps the pooled student context to the target-representation space.

For a valid cell--block pair $(i,k)$, the teacher target is the mean of the teacher token states over $T_{ik}$, followed by EMA centring and gradient stopping. If $h^{t}_{ig}$ is the teacher state of gene $g$ and $c$ is the running centre, the target is
\[
t_{ik}=\operatorname{sg}\!\left(\frac{1}{|T_{ik}|}\sum_{g\in T_{ik}}h^{t}_{ig}-c\right).
\]
The corresponding student state $s_{ik}$ is obtained by query-conditioned pooling of the visible context, and the prediction is $\widehat t_{ik}=p_{\phi}(s_{ik})$. Aggregation therefore precedes the prediction error: one graph-connected block contributes one latent target, irrespective of how many of its genes are observed. This is the defining difference from token-level JEPA, which assigns a separate target to every masked gene.

\subsection{Block-level objective}

For the set $\mathcal B$ of valid cell--block pairs in a minibatch, the objective is the mean squared distance between predicted and teacher block states,
\[
\mathcal L_{\mathrm{GB}}=\frac{1}{|\mathcal B|}\sum_{(i,k)\in\mathcal B}
\|\widehat t_{ik}-t_{ik}\|_2^2.
\]
The primary objective consists of one latent-alignment term for each accepted graph block. The graph block and aggregation operation therefore determine the supervision unit.

This construction has two consequences. First, genes from the same connected block provide redundant evidence for a shared latent target, reducing the dependence of supervision on any one sparse measurement. Second, the number of prediction terms is determined by the accepted blocks rather than by the number of detected target genes.

\subsection{Linear-attention encoder}

All student, teacher and predictor layers use linear attention. A positive feature map, $\varphi(u)=\operatorname{ELU}(u)+1$, contracts keys and values into a shared summary before each query is applied. At fixed hidden width, this factorization grows linearly with the number of observed genes\cite{katharopoulos2020linear}, enabling the encoder to retain the full observed-gene sequence.

\subsection{Cell-level embedding}

After training, the encoder is frozen and each cell is represented by pooling its gene-token states. For the 768-dimensional token matrix, we concatenate the element-wise maximum with the mean of the five largest values in each feature dimension, producing a 1,536-dimensional cell embedding. Technical-token positions are excluded from this operation. The frozen cell embedding is used for annotation, reconstruction, perturbation prediction and representation diagnostics.

\subsection{Pretraining data and optimization}

Pretraining used a scBaseCount snapshot containing records available no later than 25 February 2025\cite{youngblut2025scbasecount}. Before shuffling and graph estimation, we matched identifiable CellBench source accessions against the available scBaseCount metadata and excluded the corresponding matrices. The resulting corpus comprised 22,029,222 cells in 441 shards, and the coexpression and merged target graphs were rebuilt from these retained cells. A cell-level audit compared 86,843 available cells from the excluded standardized parent matrices with every retained cell using independent fingerprints of non-zero gene support and support plus count values. The exhaustive scan found no candidate match under either fingerprint.

We trained with an effective batch of 512 cells and analysed the representation after 20,000 optimizer steps, corresponding to 10.24 million cell presentations. This is 46.48\% of one traversal of the local collection and 2.04\% of the 502-million-cell scBaseCount resource on a presentation-equivalent basis. Supplementary Methods specify the accession exclusion and cell-fingerprint audit.

\subsection{Evaluation framework}

\method{} is the method studied throughout this work. We contextualized its representation behaviour with controls that alter the source or scale of target blocks, remove seed-local neighbourhood aggregation, replace latent block prediction with token prediction or expression decoding, or modify graph topology while retaining the encoder family. Downstream comparisons use the same embedding-extraction rule and task split within each controlled experiment. Complete variant definitions are provided in the Supplementary Information.

\subsection{Within-dataset few-shot annotation}

We followed the CellBench-LS within-dataset protocol\cite{xu2026cellbench}. For each cell type, one, five or nine labelled support cells were selected by label-stratified sampling, an SVM was fitted to frozen embeddings and all remaining labelled cells formed the held-out set. Macro-$F_1$ was the unweighted mean of class-wise $F_1$ values. We used five fixed support-set seeds; reported error bars show one standard deviation.

\subsection{Frozen-embedding reconstruction}

To measure information accessible from the representation, we trained the same multilayer perceptron on each model's frozen embedding to predict a canonical 400-gene expression vector. We evaluated cell-wise Pearson correlation, normalized root-mean-square error, overlap among the top 50 expressed genes, Reactome programme correlation\cite{gillespie2022reactome} and overlap between 15-nearest-neighbour cell-state graphs. The decoder architecture, data split and optimization budget were shared across models.

\subsection{Perturbation-response prediction}

The CellBench response model received a frozen control-cell embedding and a perturbation identity and predicted the corresponding perturbed expression profile. Adamson, Norman and Dixit were evaluated with matched Top-$k$ control examples and five fixed seeds. We quantified expression and log-fold-change error, Pearson correlation and overlap among the 50 strongest response genes. Norman additionally provided single- and combination-perturbation strata.

\subsection{Efficiency evaluation}

Embedding extraction and one-epoch partial fine-tuning were measured separately on hPancreas with batch size 8 on one NVIDIA A100-SXM4-80GB GPU using bfloat16 computation. Both \method{} and scFoundation received the same ordered cells. For partial fine-tuning, the final encoder block, leaf normalization parameters and a common two-layer classification head were trainable, yielding 26,088,205 and 26,104,077 trainable parameters, respectively. Training throughput includes forward propagation, backpropagation and the optimizer step; held-out throughput measures evaluation-mode encoding followed by the shared prediction head.

\subsection{Biological representation analyses}

For the max-token analysis, we identified the gene providing the largest token contribution in each held-out hPancreas cell and calculated its winner fraction within each annotated cell type. For programme ablation, genes from a graph-defined set were removed and the embedding displacement was compared with expression- and detection-matched random removals. Directed programme dependence was measured as the change in teacher-target prediction error after ablating a context programme, corrected by the matched-random effect.

For Norman double perturbations, each perturbation was represented by its mean displacement from control cells in the frozen embedding space. The sum of the two single-perturbation displacements was compared with the observed double-perturbation displacement, with expression-matched random gene pairs defining the null distribution.

\subsection{Representation geometry and statistics}

Effective rank and participation ratio were calculated from the eigenvalue spectrum of the centred embedding covariance. Association with sequencing depth was summarized by the absolute Pearson correlation of detected-gene count with the leading embedding axis and with embedding norm. We report these two associations as defined technical-axis diagnostics.

Few-shot and perturbation analyses used five fixed seeds unless stated otherwise. Cross-dataset intervals were obtained by nonparametric bootstrap resampling of datasets. Programme-ablation and double-perturbation effects were evaluated with paired nonparametric tests using the statistical units specified in the corresponding figure legends. Detailed metric definitions, pseudocode and complete evaluation protocols are provided in the Supplementary Information.

\section{Acknowledgements}

This work was supported by the National Key R\&D Program of China (No. 2022ZD0115100), the National Natural Science Foundation of China (No. U21A20427), the Center of Synthetic Biology and Integrated Bioengineering of Westlake University (No. WU2022A009), the ``Pioneer'' and ``Leading Goose'' R\&D Program of Zhejiang (No. 2024C01140), the Key Research and Development Program of Hangzhou (No. 2023SZD0073), the InnoHK programme and the CAAI--Ant Research Fund. We thank the Westlake University HPC Center for providing computational resources.

\section{Data availability}

The pretraining shards were derived from the public scBaseCount resource, and the evaluation datasets are available through the CellBench-LS release (\url{https://github.com/sky-Yongjie-Xu/2026-CellBench}) and their original repositories. The public repository contains the source-accession exclusion manifest, the resumable cell-fingerprint audit, compact corpus summaries and the zero-candidate audit output. The arXiv ancillary bundle contains the fixed dataset identities, benchmark protocol metadata and panel-level numerical values underlying every quantitative figure and supplementary summary. Count matrices remain available from their original public sources; the study uses public, non-identifiable single-cell data.

\section{Code availability}

The \method{} checkpoint, training and inference software, fixed graph, vocabulary and an AnnData-to-embedding interface will be released through GitHub. The repository will also document data preparation, representation extraction and reproduction of the reported analyses. Numerical source data for the figures accompany the arXiv submission.

\endgroup

\clearpage
\begingroup
\small
\bibliographystyle{unsrtnat}
\bibliography{references}
\endgroup

\clearpage
\appendix
\renewcommand{\thetable}{S\arabic{table}}
\setcounter{table}{0}
\renewcommand{\thealgorithm}{S\arabic{algorithm}}
\setcounter{algorithm}{0}
\section*{Supplementary Information}

\section*{Supplementary Methods}

\subsection*{Study design and notation}

We evaluated \method{} as a frozen embedding model and used the same saved representation for annotation, clustering, reconstruction and perturbation prediction. Let $x_i\in\mathbb R^G$ denote the canonical expression vector of cell $i$, $y_i\in\{1,\ldots,C\}$ its cell-type label and $z_i=f(x_i)\in\mathbb R^d$ the frozen model embedding. All supervised heads were trained after freezing $f$. We used fixed split identities and the same random seeds for every model within a task. The reported benchmark models were UMAP, scVI, CellPLM, Geneformer, LangCell, scGPT, scMulan, scFoundation, Nicheformer and \method{} where outputs were available.

\subsection*{Pretraining corpus, sampling and exposure accounting}

The pretraining shards were prepared from a scBaseCount snapshot containing records available no later than 25 February 2025. scBaseCount is an AI-agent-curated and uniformly processed single-cell repository\cite{youngblut2025scbasecount}; the current source repository contains more than 502 million cells across organisms, tissues and studies. The present model was trained on a curated shard collection rather than for one full pass over the complete repository.

The pretraining collection contained 441 CSR sparse matrices with 19,264 columns. The first 440 shards contained 50,000 rows and the final shard contained 29,222 rows, for a total of
\[
N_{\rm local}=440\times50{,}000+29{,}222=22{,}029{,}222
\]
cells. Shard order and row order within each shard were shuffled. Shards were partitioned across eight distributed ranks and then across eight data-loader workers per rank, so each row was assigned to one rank--worker pair during a traversal.

Pretraining used eight GPUs, a local batch of eight cells per GPU and eight gradient-accumulation batches. If $B_{\rm local}$ is the local batch, $D$ the number of distributed workers and $A$ the accumulation factor, one optimizer step represents
\[
 B_{\rm eff}=B_{\rm local}DA=8\times8\times8=512
\]
cell presentations. The representation analysed in this study was obtained at optimizer step $s=20{,}000$, giving
\[
 N_{\rm present}=sB_{\rm eff}=10{,}240{,}000
\]
cell presentations. One complete local pass requires approximately
\[
N_{\rm local}/B_{\rm eff}=43{,}025.8
\]
optimizer steps. The analysed representation covers 46.48\% of the local rows before completion of the first traversal. Sortish batching assigns each row once within a traversal, giving $N_{\rm unique}\simeq N_{\rm present}$ up to terminal incomplete worker batches. Relative to the 502-million-cell resource, 10.24 million corresponds to 2.04\% on a presentation-equivalent scale.

\paragraph{Context relative to reported foundation-model corpora.}
The 10.24 million presentations are lower than the published pretraining-corpus counts for Geneformer (approximately 30 million cells), scGPT (more than 33 million), scFoundation (more than 50 million) and CellFM (102,304,686 cells)\cite{theodoris2023geneformer,cui2024scgpt,hao2024scfoundation,zeng2025cellfm}. Table~\ref{tab:reported_training_scale} reports these quantities without converting them into nominal epochs.

\begin{table}[ht]
\centering
\suppTableStyle
\caption{Cell scale used to contextualize pretraining exposure. The \method{} value is the number of cell presentations used in this study; comparator values are published corpus sizes.}
\label{tab:reported_training_scale}
\begin{tabular}{lll}
\toprule
Model & Reported scale & Quantity represented \\
\midrule
\biomrow\textbf{\method{}} & 10.24 million & Cell presentations \\
Geneformer & $\sim$30 million & Pretraining corpus cells \\
scGPT & $>$33 million & Pretraining corpus cells \\
scFoundation & $>$50 million & Pretraining corpus cells \\
CellFM & 102.3 million & Pretraining corpus cells \\
\bottomrule
\end{tabular}
\end{table}

These quantities place the training exposure in context: the reported representation emerged after fewer cell presentations than the published corpus size of each comparator. Because the models differ in architecture, vocabulary and preprocessing, Table~\ref{tab:reported_training_scale} is an exposure comparison rather than a fitted scaling law.

\paragraph{Evaluation-overlap audit.}
Pretraining and graph construction used unlabeled expression matrices. We constructed the pretraining corpus by excluding identifiable benchmark source records before shuffling, and audited it at both source-record and cell-matrix levels.

At the source-record level, we assembled a manifest of the studies underlying the CellBench annotation, reconstruction and perturbation datasets and matched their accessions against the available scBaseCount metadata. This identified 242 experiment records associated with one parent study, PRJEB51634. Of these, 102 standardized parent matrices were present locally and contained 86,843 cells. We removed these matrices before the two-pass global shuffle. The remaining 12,822 non-empty inputs yielded 22,029,222 cells in 441 CSR shards, and both the corpus-derived coexpression graph and the merged STRING--coexpression target graph were rebuilt from this retained corpus.

At the cell-matrix level, every excluded cell and every retained cell was represented in the ordered 19,264-gene vocabulary. For a sparse row with ordered non-zero gene indices $J_i$ and corresponding values $X_i$, we computed two domain-separated 128-bit fingerprints,
\[
H_{\rm supp}(i)=\operatorname{BLAKE2b}_{128}
\!\left(\operatorname{LE32}(J_i);\,\mathrm{BMJ\!\mathchar`-support\!\mathchar`-v1}\right)
\]
and
\[
H_{\rm count}(i)=\operatorname{BLAKE2b}_{128}
\!\left(\operatorname{LE32}(J_i)\,\Vert\,\operatorname{LEf32}(X_i);\,
\mathrm{BMJ\!\mathchar`-count\!\mathchar`-v1}\right),
\]
where $\operatorname{LE32}$ and $\operatorname{LEf32}$ denote canonical little-endian int32 and float32 encodings. The support fingerprint is invariant to value normalization when the non-zero support is unchanged; the count fingerprint additionally tests equality of the standardized values. Fingerprints were used only as a screening index. Any candidate digest match was reloaded from both sparse matrices and verified by exact comparison of the index and value arrays.

The 86,843 excluded query cells had 86,843 unique support fingerprints and 86,843 unique count fingerprints. The exhaustive retained-corpus scan covered 22,029,222 cells, 441 shards and 82,283,764,256 non-zero entries. It produced zero support candidates, zero count candidates and zero verified matches against the available standardized PRJEB51634 parent matrices.

\begin{table}[ht]
\centering
\suppTableStyle
\caption{Accession exclusion and cell-matrix fingerprint audit for the decontaminated retraining corpus. A candidate fingerprint would have been verified by exact sparse-row comparison.}
\label{tab:cellbench_overlap_audit}
\resizebox{\linewidth}{!}{%
\begin{tabular}{lrrl}
\toprule
Audit stage & Records or cells & Retained-corpus candidates & Outcome \\
\midrule
Metadata accession match & 242 experiment records & -- & Matched to PRJEB51634 \\
Locally available parent matrices & 102 matrices; 86,843 cells & -- & Excluded before shuffling \\
Non-zero support fingerprint & 86,843 query cells & 0 of 22,029,222 cells & 0 verified support matches \\
Support-plus-count fingerprint & 86,843 query cells & 0 of 22,029,222 cells & 0 verified count matches \\
Corpus-derived graph & 22,029,222 retained cells & -- & Rebuilt after exclusion \\
\bottomrule
\end{tabular}
}
\end{table}

The resulting corpus provides source-level exclusion for accession-resolved inputs and exact-row exclusion for the 86,843 available standardized parent cells. Annotation is reported uniformly as within-dataset frozen-probe evaluation for \method{} and comparator models.

\paragraph{Temporal audit against the pretraining cutoff.}
The CellBench-LS manuscript was posted in 2026, after the 25 February 2025 scBaseCount snapshot cutoff. We traced its constituent datasets, together with the additional Dixit perturbation dataset, to their original sources\cite{xu2026cellbench}. Source-publication years ranged from 2017 to 2024 for the general-purpose datasets, 2021 for the batch-correction datasets and 2016--2019 for the perturbation datasets (Table~\ref{tab:cellbench_temporal_audit}). These dates place the benchmark studies within the period covered by the snapshot and motivate the accession and cell-fingerprint audit above.

\begin{table}[ht]
\centering
\suppTableStyle
\caption{Benchmark source years relative to the 25 February 2025 pretraining snapshot. Years denote the source publications reported by CellBench-LS or, for the additional Dixit dataset, its original study.}
\label{tab:cellbench_temporal_audit}
\resizebox{\linewidth}{!}{%
\begin{tabular}{llll}
\toprule
Evaluation family & Dataset & Source publication year & Relation to snapshot \\
\midrule
Annotation/clustering & PBMC12k; hPancreas; MS; Zheng68k & 2020; 2022; 2019; 2017 & By cutoff \\
Annotation/clustering & cortex; COVID; liver; lung; immune & 2023; 2024; 2021; 2020; 2022 & By cutoff \\
Batch correction & UC-EPI; UC-IMM & 2021; 2021 & By cutoff \\
Perturbation & Adamson; Norman & 2016; 2019 & By cutoff \\
Additional perturbation & Dixit & 2016 & By cutoff \\
\bottomrule
\end{tabular}
}
\end{table}

\subsection*{CellBench-LS datasets and preprocessing}

Annotation and clustering used the nine CellBench-LS datasets PBMC12k, hPancreas, cortex, multiple sclerosis (MS), liver, Zheng68k, lung, COVID and immune. Reconstruction used hPancreas, cortex, MS, liver, Zheng68k, lung, COVID and immune. Perturbation prediction used Adamson, Norman and Dixit; combination perturbations occur in Norman. Raw count matrices were retained where required. Gene identifiers were mapped to each model's published vocabulary, duplicated identifiers were resolved before inference and cells lacking a valid benchmark label were removed.

Each model retained its documented tokenization and value transformation. \method{} mapped sparse raw counts to the 19,264-gene vocabulary and applied $v_{ig}=\log(1+10^4c_{ig}/\sum_{g'}c_{ig'})$. Two scFoundation-style control values represented a fixed high-resolution value of 4 and $\log_{10}$ library size. They were visible context entries for the student and teacher but were excluded from graph-target eligibility. Frozen \method{} embeddings excluded these positions and concatenated the element-wise maximum gene-token state with the mean of the five largest gene-token states per feature, producing a 1,536-dimensional embedding. The same pooling and control-position policy was used for matched objective controls.

\subsection*{Linear-attention architecture}

\method{} used a 19,264-gene vocabulary, hidden width $d=768$, twelve encoder layers and twelve heads per layer. All encoder layers used flash linear attention. The prediction head used four linear-attention layers, twelve heads and dropout $0.05$. The student was fully trainable. The teacher shared the encoder architecture and was updated only by EMA with momentum annealed from $0.996$ to $0.9997$.

For one head, let $Q,K,V\in\mathbb R^{n\times d_h}$ and $\varphi(u)=\operatorname{ELU}(u)+1$. Linear attention evaluates
\[
 y_i=
 \frac{\varphi(q_i)^\top\left[\sum_j\varphi(k_j)v_j^\top\right]}
 {\varphi(q_i)^\top\left[\sum_j\varphi(k_j)\right]+10^{-6}},
\]
after setting padded keys and values to zero. The numerator and normalizer are contracted once rather than materializing an $n\times n$ token-affinity matrix. With $h$ heads and $d_h=d/h$, the attention-core complexity is
\[
\mathcal O\!\left(hnd_h^2\right)
=\mathcal O\!\left(\frac{nd^2}{h}\right),
\]
compared with $\mathcal O(n^2d)$ for dense softmax attention. This asymptotic analysis covers the attention core at fixed width; the measured end-to-end throughput additionally includes tokenization, feed-forward layers and input/output transfer.

\subsection*{Biological graph and target blocks}

The implemented target graph was restricted to the shared gene vocabulary $\mathcal V$, with $|\mathcal V|=19{,}264$. It used two sources.

\paragraph{STRING protein-association adjacency.}
STRING v12 human links were mapped to the vocabulary by preferred symbol and unambiguous aliases\cite{szklarczyk2023string}. Links with combined score below 700 and self-edges were removed. For each row, the 64 highest-scoring mapped neighbours were retained and the result was symmetrized. This graph mapped 17,892 STRING proteins and contained 356,772 nonzero adjacency entries.

\paragraph{Corpus coexpression adjacency.}
All 22,029,222 pretraining cells were transformed as
\[
x_{ig}'=\log\!\left(1+10^4x_{ig}/\sum_hx_{ih}\right).
\]
Centred signed CountSketch projections with 256 dimensions approximated the absolute Pearson association between every gene pair. Symmetric top-32 and top-64 coexpression graphs contained 1,071,500 and 2,072,226 nonzero entries, respectively. The top-32 graph was used to construct target blocks and the top-64 graph defined the query-local context neighbourhood. Coexpression was therefore estimated entirely from unlabeled expression matrices.

The STRING and corpus-coexpression adjacencies were binarized, combined by union and truncated row-wise to 64 neighbours:
\[
A=\operatorname{RowFirst}_{64}
\left(\mathbf 1[A^{\rm STRING}_{64}>0]\lor\mathbf 1[A^{\rm coexp}_{32}>0]\right).
\]
Deterministic row-wise truncation after the union gives $A$ a directed neighbourhood-table representation. It contains 1,002,591 entries and has mean out-degree 52.04. Its edge sources are STRING protein associations and corpus coexpression.

\paragraph{Operational target-block definition.}
In this work a graph-connected gene block is a \emph{candidate target set} under $A$. Biological programme and pathway are reserved for independently supported gene sets. For cell $i$, let $\mathcal O_i$ be its observed genes. For each of $K=4$ targets, we draw a seed
\[
r_{ik}\sim\operatorname{Uniform}(\mathcal O_i)
\]
and an integer requested size
$L_{ik}\sim\operatorname{Uniform}\{512,\ldots,2000\}$.
Breadth-first expansion over outgoing neighbour lists gives
\[
B_{ik}=\operatorname{BFS}(A,r_{ik},L_{ik}).
\]
The cell-specific target support is the observed intersection
\[
T_{ik}=B_{ik}\cap\mathcal O_i.
\]
The teacher block state is averaged over $T_{ik}$. The query identifier is the BFS seed $r_{ik}$ and combines its learned position embedding with a learned mask vector.

The student context is
\[
\mathcal C_i=\mathcal O_i\setminus\bigcup_{k=1}^{K}T_{ik}.
\]
The student input contains the observed complement of the target union. On the student side, query-conditioned pooling restricts context candidates to
\[
\mathcal C_{ik}^{\rm DMT}
=\mathcal C_i\cap\mathcal N^{\rm coexp}_{64}(r_{ik})
\]
and the sampler accepts blocks with a non-empty seed-local intersection. The teacher receives the full observed gene set.

\paragraph{Disjoint target and context construction.}
Target masks are constructed before the student encoder is evaluated. For every seed, the sampler reserves one observed coexpression neighbour in the student context and removes this reserved anchor from all target masks. A target group is accepted when each block retains seed-local context; sampling continues within the same construction step until this condition is met. Consequently, $\mathcal C_i\cap(\bigcup_kT_{ik})=\varnothing$, and every student block summary is formed from seed-local non-target genes.

\begin{table}[ht]
\centering
\suppTableStyle
\caption{Target--context construction used in pretraining.}
\label{tab:target_context_construction}
\begin{tabular}{ll}
\toprule
Condition & Pretraining behaviour \\
\midrule
Student context & Observed complement of the target union \\
Target--context relation & Disjoint supports \\
Seed-local context & At least one observed coexpression neighbour per block \\
Empty seed-neighbourhood context & Continue target-group sampling \\
Query identity & BFS seed gene \\
\bottomrule
\end{tabular}
\end{table}

Random-block controls sampled candidate gene sets uniformly from the vocabulary and used the same aggregate target construction. Graph-rewiring and block-scale controls retained the block-level objective while changing the distribution of graph-connected targets. The exact size range, neighbourhood policy and topology change for each variant are reported with the corresponding ablation results.

\subsection*{Block aggregation and \method{} loss}

\paragraph{Relation to established JEPA objectives.}
The general JEPA principle predicts a target state in representation space from its context\cite{lecun2022jepa}. In I-JEPA, a context encoder and predictor estimate a collection of teacher patch embeddings for masked image regions\cite{assran2023ijepa}; V-JEPA applies the same feature-prediction principle to masked spatiotemporal regions\cite{bardes2024vjepa}. Denoting the patch or token targets in block $T_k$ by $\{h_i^t:i\in T_k\}$, the corresponding token-level objective has the generic form
\[
  \mathcal L_{\rm token}
  =
  \frac{1}{\sum_k|T_k|}
  \sum_k\sum_{i\in T_k}
  \ell\!\left(\widehat h_i,\operatorname{sg}(h_i^t)\right).
\]
\method{} retains the teacher--student principle and defines the supervision unit as an aggregate graph block. It first maps each graph-defined gene set to one teacher block state
\[
  t_k=\operatorname{sg}\!\left[
  \operatorname{Agg}\{h_i^t:i\in T_k\}-c\right],
\]
and then aligns one predicted state $\widehat t_k$ with $t_k$. Consequently, the implemented loss contains $K$ aggregate block-level prediction terms.

The student and EMA teacher produced token matrices
\[
 H_i^{s}=f_\theta(x_{i,\mathcal C}),\qquad
 H_i^{t}=f_{\bar\theta}(x_i),\qquad
 \bar\theta\leftarrow\mu\bar\theta+(1-\mu)\theta .
\]
For cell-specific block support $T_{ik}$ with BFS seed-gene index $r_{ik}$, the query was
\[
 q_{ik}=e(r_{ik})+m,
\]
where $e(r_{ik})$ is the learned position embedding of the seed gene and $m$ is a learned mask embedding. This query identifies the graph expansion that defines the block. The DMT aggregator restricts context candidates to $\mathcal C_{ik}=\mathcal C_i\cap\mathcal N_{64}^{\rm coexp}(r_{ik})$. Accepted blocks therefore aggregate a non-empty seed-local context:
\[
 s_{ik}=\sum_{j\in\mathcal C_{ik}}a_{ikj}h^s_{ij},\qquad
 a_{ikj}=
 \frac{\exp\{\cos(q_{ik},h^s_{ij})/\tau\}}
 {\sum_{\ell\in\mathcal C_{ik}}\exp\{\cos(q_{ik},h^s_{i\ell})/\tau\}}.
\]
The teacher block state was aggregated before any prediction error was formed,
\[
  t_{ik}=\operatorname{sg}\!\left[
  |T_{ik}|^{-1}\sum_{j\in T_{ik}}h^t_{ij}-c
  \right],\qquad
  \widehat t_{ik}=p_\phi(s_{ik}),
\]
where $c$ is an EMA centre and $\operatorname{sg}$ stops gradients. For the valid cell--block index set $\mathcal B$, alignment was
\[
\mathcal L_{\rm align}=
\frac{\lambda_2}{|\mathcal B|}
\sum_{(i,k)\in\mathcal B}\|\widehat t_{ik}-t_{ik}\|_2^2+
\frac{\lambda_s}{|\mathcal B|}
\sum_{(i,k)\in\mathcal B}
\left\|
\frac{\widehat t_{ik}}{\|\widehat t_{ik}\|_2}-
\frac{t_{ik}}{\|t_{ik}\|_2}
\right\|_2^2 .
\]
We used $\lambda_2=1$ and $\lambda_s=0$, giving the optimized objective
\[
\boxed{
\mathcal L_{\rm GB\text{-}JEPA}=\mathcal L_{\rm align}}.
\]
Each accepted block contributes one term to this objective. The token-IJEPA control optimized
\[
\mathcal L_{\rm token}=
\frac{1}{|\mathcal T|}
\sum_{(i,j)\in\mathcal T}
\ell\!\left(\widehat h_{ij},\operatorname{sg}(h^t_{ij})\right).
\]
The decoder-only control used expression reconstruction as its sole pretraining objective.

\begin{algorithm}[ht]
\caption{\method{} pretraining}
\label{alg:gbjepa}
\begin{algorithmic}[1]
\Require Minibatch $X$; gene graph $\mathcal G$; student $f_\theta$; teacher $f_{\bar\theta}$; predictor $p_\phi$; EMA rate $\mu$
\For{each minibatch $X$}
  \State $(\mathcal T,\mathcal R)\gets\Call{SampleBFSBlocks}{X,\mathcal G,512,2000,K=4}$
  \State $(\mathcal T,\mathcal R)\gets\Call{RequireSeedLocalContext}{\mathcal T,\mathcal R,\mathcal N^{\rm coexp}_{64}}$
  \State $\mathcal C\gets\mathcal O(X)\setminus\bigcup_{T\in\mathcal T}T$
  \State $H^s\gets f_\theta(X_{\mathcal C})$
  \State $H^t\gets\Call{StopGradient}{f_{\bar\theta}(X)}$
  \State $\widehat{\mathcal Z}\gets[\,]$; $\mathcal Z\gets[\,]$
  \For{each valid cell--block support and seed $(T_{ik},r_{ik})\in(\mathcal T,\mathcal R)$}
    \State $q_{ik}\gets e(r_{ik})+m$
    \State $s_{ik}\gets\Call{ContextPool}{H^s_i,q_{ik},\mathcal N^{\rm coexp}_{64}(r_{ik})}$
    \State $t_{ik}\gets\Call{MeanPool}{H^t_{i,T_{ik}}}-c$
    \State $\widehat{\mathcal Z}\gets\widehat{\mathcal Z}\mathbin{\|}p_\phi(s_{ik})$
    \State $\mathcal Z\gets\mathcal Z\mathbin{\|}\Call{StopGradient}{t_{ik}}$
  \EndFor
  \State $\mathcal L\gets\Call{MSE}{\widehat{\mathcal Z},\mathcal Z}$
  \State $(\theta,\phi)\gets\Call{OptimizerStep}{\mathcal L,\theta,\phi}$
  \State $\bar\theta\gets\mu\bar\theta+(1-\mu)\theta$
  \State $c\gets\Call{UpdateEMACentre}{c,\mathcal Z}$
\EndFor
\end{algorithmic}
\end{algorithm}

\subsection*{Objective curves and target similarity}

Raw objectives have different units; therefore, the curve in Fig.~2a was normalized within run by the median loss during the first 400 optimization steps:
\[
\widetilde{\mathcal L}_r(s)=
\frac{\mathcal L_r(s)}
{\operatorname{median}_{u\leq400}\mathcal L_r(u)}.
\]
We plotted a centred rolling median and summarized local dispersion with the rolling standard deviation of residuals. At every 800-step training snapshot, we quantified target agreement with orthogonal-Procrustes cosine similarity. For centred prediction and teacher matrices $P,T\in\mathbb R^{n\times d}$,
\[
R^\star=\arg\min_{R^\top R=I}\|PR-T\|_F^2,\qquad
S_{\rm Proc}=\frac{1}{n}\sum_{i=1}^{n}
\cos(P_iR^\star,T_i).
\]
This rotation-invariant measure compares the geometry of the two latent sets independently of their coordinate bases. It was evaluated at the native supervision unit: blocks for \method{} and genes for token-IJEPA.

\subsection*{Embedding geometry and depth association}

For centred embedding matrix $Z\in\mathbb R^{N\times d}$ with covariance eigenvalues $\lambda_r$, define $p_r=\lambda_r/\sum_j\lambda_j$. Effective rank and participation ratio were
\[
r_{\rm eff}=\exp\!\left(-\sum_rp_r\log p_r\right),\qquad
r_{\rm PR}=\frac{(\sum_r\lambda_r)^2}{\sum_r\lambda_r^2}.
\]
Fig.~2b evaluated effective rank every 800 steps on hPancreas. Fig.~3c evaluated both statistics on hPancreas and cortex.

For each cell we retained total counts $d_i^{\rm total}$ and the number of detected genes $d_i^{\rm nnz}$. Let $u_i$ be the score along the leading embedding axis and $n_i=\|z_i\|_2$. The four technical associations were
\[
|r(u,d^{\rm total})|,\quad |r(u,d^{\rm nnz})|,\quad
|r(n,d^{\rm total})|,\quad |r(n,d^{\rm nnz})|,
\]
where $r$ is Pearson correlation. The main control figure displays the detected-gene associations. The failure-mode summary reports their mean over the retained diagnostics and uses dataset values as the uncertainty unit.

These diagnostics quantify association along the leading embedding axis and embedding norm. Random-block, token-IJEPA and decoder-only provide comparisons among complete pretraining formulations; their interpretation is therefore confined to the tested objective families.

\subsection*{Few-shot annotation}

For each dataset, top-$k$ annotation used $k\in\{1,5,9\}$ labelled cells per class as the support set. Support indices were sampled independently for each class, and all remaining labelled cells formed the held-out evaluation set. Splitting was label-stratified. A support-vector machine was trained on frozen embeddings with identical hyperparameters for all models. We used five fixed seeds, $42$--$46$.

For class $c$, precision, recall and $F_1$ were
\[
P_c=\frac{\mathrm{TP}_c}{\mathrm{TP}_c+\mathrm{FP}_c},\qquad
R_c=\frac{\mathrm{TP}_c}{\mathrm{TP}_c+\mathrm{FN}_c},\qquad
F_{1,c}=\frac{2P_cR_c}{P_c+R_c}.
\]
Accuracy and macro-$F_1$ were
\[
\operatorname{Acc}=\frac{1}{N}\sum_i\mathbf 1[\widehat y_i=y_i],
\qquad
F_1^{\rm macro}=\frac{1}{C}\sum_{c=1}^{C}F_{1,c}.
\]
Figures~2d and 3b use Top-5 macro-$F_1$ on hPancreas and cortex. Points are means over five seeds and error bars are one standard deviation unless stated otherwise.

\subsection*{Matched efficiency benchmark}

We used two complementary timing protocols. \method{} and scFoundation were evaluated on the same NVIDIA A100-SXM4-80GB GPU with PyTorch 2.7.1, CUDA 11.8 and bfloat16 computation. The same ordered raw-count cells and 19,264-gene vocabulary were used. Gene mapping and disk input/output were completed before each timed interval.

\paragraph{Encoder-only extraction.}
We prepared 2,048 cells from hPancreas and used batch size 8, eight warm-up iterations and 40 synchronized timed iterations. Each sequence retained its full observed-gene set. hPancreas contained 14,818 available cells and had observed-gene lengths of 69--1,659 (median, 319; mean, 386.8; 95th percentile, 931). \method{} used its 1,536-dimensional maximum-plus-top-five-mean embedding; scFoundation used its documented four-part 3,072-dimensional control-token/max/mean representation. Each model was evaluated through its native frozen embedding. If $N$ cells are encoded between synchronized times $t_0$ and $t_1$, throughput is
\[
  R_{\rm emb}=\frac{N}{t_1-t_0}.
\]

\begin{table}[ht]
\centering
\suppTableStyle
\caption{Encoder-only embedding throughput. Values are cells s$^{-1}$; the final column is the \method{}/scFoundation ratio.}
\begin{tabular}{lrrrr}
\toprule
Dataset & Batch & \method{} & scFoundation & Speed-up \\
\midrule
\biomrow hPancreas & 8 & 678.26 & 353.35 & 1.92$\times$ \\
\bottomrule
\end{tabular}
\end{table}

\paragraph{One-epoch partial fine-tuning.}
We used 1,279 hPancreas training cells, 427 validation cells and 427 test cells, with batch size 8 and one epoch. Both models received the same 19,264 genes plus two scFoundation-style control values and formed the same four-part pooled vector (last control, second-last control, gene-wise maximum and gene-wise mean; 3,072 dimensions). Each used the identical head
\[
\operatorname{LayerNorm}\rightarrow
\operatorname{Linear}(3072,6144)\rightarrow
\operatorname{GELU}\rightarrow
\operatorname{Linear}(6144,C).
\]
The head learning rate was $10^{-3}$, the encoder learning rate was $5\times10^{-5}$, weight decay was $0.01$ and AdamW coefficients were $(0.9,0.999)$. The trainable set comprised the final encoder block, direct parameters of leaf normalization layers and the common head. The resulting encoder-trainable counts were 7,121,664 for \method{} and 7,137,536 for scFoundation; total trainable counts including the common head were 26,088,205 and 26,104,077.

Training throughput includes forward propagation, backward propagation and the optimizer step,
\[
  R_{\rm train}=\frac{N_{\rm global}}{t_{\rm step}},
\]
whereas held-out throughput is
\[
R_{\rm heldout}=
\frac{N_{\rm valid}+N_{\rm test}}{t_{\rm valid}+t_{\rm test}}.
\]
The held-out loop computed evaluation-mode encoder embeddings followed by the identical head. Its 3.76-fold ratio measures task-time embedding throughput; the isolated encoder-only protocol yielded a 1.92-fold ratio.

\begin{table}[ht]
\centering
\suppWideTableStyle
\caption{One-epoch hPancreas timing. One A100 GPU, bfloat16 and batch size 8 were used.}
\resizebox{\linewidth}{!}{%
\begin{tabular}{lrrrrrrr}
\toprule
Model & Train (s) & Valid (s) & Test (s) & Train cells s$^{-1}$ & Held-out cells s$^{-1}$ & Trainable & Peak GiB \\
\midrule
\biomrow\textbf{\method{}} & 52.48 & 6.87 & 6.73 & 24.37 & 62.78 & 26,088,205 & 27.98 \\
scFoundation & 301.49 & 25.93 & 25.26 & 4.24 & 16.68 & 26,104,077 & 24.59 \\
\midrule
\rankrow \method{}/scFoundation & --- & --- & --- & \textbf{5.75$\times$} & \textbf{3.76$\times$} & --- & --- \\
\bottomrule
\end{tabular}}
\end{table}

The train--validation--test epoch took 66.08\,s for \method{} and 352.68\,s for scFoundation, corresponding to a 5.34-fold wall-clock speed-up.

\subsection*{Clustering}

We clustered frozen embeddings with $k$-means using the true number of benchmark classes, while withholding labels during clustering. Cluster accuracy used the optimal Hungarian assignment $\pi$:
\[
\operatorname{ACC}_{\rm clust}=
\max_{\pi}\frac{1}{N}\sum_i\mathbf 1[y_i=\pi(c_i)].
\]
Normalized mutual information (NMI), adjusted Rand index (ARI) and mean silhouette width (ASW) were computed from the same assignment. With contingency counts $n_{ab}$, row sums $n_{a\cdot}$, column sums $n_{\cdot b}$ and $N=\sum_{ab}n_{ab}$,
\[
\operatorname{NMI}=
\frac{2\sum_{ab}\frac{n_{ab}}{N}
\log\frac{Nn_{ab}}{n_{a\cdot}n_{\cdot b}}}
{H(Y)+H(C)}.
\]
ARI used the standard chance-corrected pair-count statistic,
\[
\operatorname{ARI}=
\frac{
\sum_{ab}\binom{n_{ab}}{2}-
\frac{\sum_a\binom{n_{a\cdot}}{2}
\sum_b\binom{n_{\cdot b}}{2}}{\binom N2}
}{
\frac12\!\left[
\sum_a\binom{n_{a\cdot}}{2}+
\sum_b\binom{n_{\cdot b}}{2}\right]-
\frac{\sum_a\binom{n_{a\cdot}}{2}
\sum_b\binom{n_{\cdot b}}{2}}{\binom N2}
}.
\]
For cell $i$, silhouette width was
$s_i=(b_i-a_i)/\max(a_i,b_i)$, where $a_i$ is its mean within-cluster distance and $b_i$ the smallest mean distance to another cluster. CellBench AvgBio is
\[
\operatorname{AvgBio}=\frac{\operatorname{NMI}+\operatorname{ARI}+\operatorname{ASW}}{3}.
\]

\subsection*{Frozen-embedding reconstruction}

Each model embedding was passed to the same $512$--$1024$--$512$ multilayer perceptron with identical optimization and early-stopping rules. We predicted a 400-gene canonical expression vector. Training budgets were 100, 500, 700 and 900 cells per class, capped by class availability. Classwise rare-safe stratification preserved every evaluable class in the held-out set. Five seeds, $42$--$46$, were used.

For truth $x_i$ and prediction $\widehat x_i$, the cell-wise Pearson score was
\[
\rho_{\rm cell}=\frac{1}{N}\sum_i
\operatorname{corr}(x_i,\widehat x_i).
\]
Normalized RMSE was
\[
\operatorname{NRMSE}=
\sqrt{\frac{\sum_i\|\widehat x_i-x_i\|_2^2}
{\sum_i\|x_i-\overline x_{\rm train}\|_2^2}}.
\]
For $S_{50}(x)$, the indices of the 50 largest expression values,
\[
\operatorname{Top50}=\frac{1}{50N}\sum_i
|S_{50}(x_i)\cap S_{50}(\widehat x_i)|.
\]
For Reactome program $m$ with gene set $G_m$, we formed
$q_{im}=|G_m|^{-1}\sum_{g\in G_m}x_{ig}$ and its predicted analogue
$\widehat q_{im}$. Program recovery was the median over programs of
$\operatorname{corr}(q_{\cdot m},\widehat q_{\cdot m})$.

Cell-state geometry was measured by $k$-nearest-neighbour overlap at $k=15$:
\[
\operatorname{kNN@15}=
\frac{1}{15N}\sum_i
|\mathcal N_{15}(x_i)\cap\mathcal N_{15}(\widehat x_i)|.
\]
Large points in Fig.~4b--d are means across eight datasets. Their 95\% intervals were obtained from 4,000 nonparametric bootstrap resamples of datasets. Small symbols are individual datasets.

For the hPancreas marker-program matrix, we selected the six most abundant cell types and the five strongest positive marker genes per type from the training partition only. Each gene was standardized with its training mean and standard deviation. The displayed entry for cell type $a$ and marker program $b$ is the held-out mean standardized score of program $b$ among cells of type $a$. No test label or test expression value entered marker selection.

\subsection*{Perturbation-response prediction}

The CellBench response predictor received a frozen control-cell embedding and a perturbation identity. Top-$k$ denotes $k\in\{1,5,9\}$ control examples. We used the benchmark perturbation splits and five seeds. Let $\mu_{pg}$ and $\mu_{0g}$ be mean expression of gene $g$ under perturbation $p$ and control, and define the observed and predicted log-fold changes
\[
\Delta_{pg}=\mu_{pg}-\mu_{0g},\qquad
\widehat\Delta_{pg}=\widehat\mu_{pg}-\mu_{0g}.
\]
The reported error was
\[
\operatorname{MSE\text{-}LFC}=
\frac{1}{|\mathcal P|G}
\sum_{p\in\mathcal P}\sum_{g=1}^{G}
(\widehat\Delta_{pg}-\Delta_{pg})^2.
\]
Pearson-LFC was the Pearson correlation after concatenating
$\{\Delta_{pg}\}$ and $\{\widehat\Delta_{pg}\}$ within each evaluation category. Categories were All, unseen single perturbations and unseen combinations; the combination category is defined only for Norman. Response-gene recovery was
\[
\operatorname{Top50}_{\rm response}=
\frac{1}{50|\mathcal P|}
\sum_{p\in\mathcal P}
|S_{50}(|\Delta_p|)\cap S_{50}(|\widehat\Delta_p|)|.
\]
Fig.~5b,d report mean $\pm$ standard deviation over five seeds. Fig.~5e shows one pre-specified Adamson Top-5 condition. Genes were selected by the magnitude of the observed response, and the colour scale displays signed response divided by the corresponding observed response magnitude, clipped symmetrically to $[-1.5,1.5]$.

\subsection*{Biological-programme and genetic-interaction analysis}

All analyses in Fig.~6 operate on frozen embeddings and token states. Cell-type and perturbation labels organize the analysis after embedding extraction.

\paragraph{Max-token cell-type programmes.}
Let $u_{ig}$ denote the contribution of gene $g$ to the pooled representation of cell $i$ under the fixed extraction rule. For cell type $c$ with held-out cell set $\mathcal I_c$, the max-token winner fraction is
\[
 A_{cg}=\frac{1}{|\mathcal I_c|}
 \sum_{i\in\mathcal I_c}
 \mathbf 1\!\left[g=\arg\max_{j\in\mathcal O(x_i)}u_{ij}\right].
\]
Genes were ordered by their largest winner fraction and the displayed heat map retained canonical endocrine, exocrine, stromal and immune examples. This analysis recovered, among others, \textit{GCG}, \textit{INS}, \textit{SST}, \textit{PPY}, \textit{GHRL} and \textit{REG1A} in their expected hPancreas cell populations.

\paragraph{Graph-programme ablation.}
For a graph block $B$ selected to represent an independently annotated cell programme, its cell-type score is
\[
 A_{cB}=\frac{1}{|\mathcal I_c|}
 \sum_{i\in\mathcal I_c}
 \frac{1}{|B\cap\mathcal O(x_i)|}
 \sum_{g\in B\cap\mathcal O(x_i)}u_{ig}.
\]
Scores are computed independently in every seed before aggregation.

For cell $i$, let $z_i$ be its unmodified embedding and $z_i^{(-B)}$ the embedding after genes in $B$ are removed. The mean displacement of cell type $c$ is
\[
D_{cB}=\frac{1}{|\mathcal I_c|}\sum_{i\in\mathcal I_c}
\left\|z_i-z_i^{(-B)}\right\|_2.
\]
Each of the eight programmes was paired with one random gene set matched for set size, mean expression and detection frequency. If $\widetilde D_{cB}$ is the mean displacement under this matched null, the excess programme effect is
\[
E_{cB}=D_{cB}-\widetilde D_{cB}.
\]
For each programme, the annotated cell identity corresponding to the programme was designated the matched identity before model inspection; all remaining identities formed the comparison group. We sampled 24 cells from each of eight identities, giving 192 unique cells. The plotted matched value for programme $B$ is the mean excess displacement over its 24 corresponding cells; the comparison value is the mean over the 168 cells from the other seven identities. The eight programme-level matched and comparison means were the statistical pairs. Their grand means were 5.34 and 0.064, respectively (two-sided paired Wilcoxon signed-rank test, $n=8$ programmes, statistic $W=0$, nominal $P=0.0078125$). This was the pre-specified aggregate hypothesis for Fig.~6b.

\paragraph{Directed programme dependencies.}
Let $\varepsilon_b(x)$ be the teacher-target prediction error for target programme $b$ in cell $x$. After ablating context programme $a$, the directed effect was
\[
I_{a\rightarrow b}=
\mathbb E_x\!\left[\varepsilon_b(x^{(-a)})-\varepsilon_b(x)\right]
-
\mathbb E_x\!\left[\varepsilon_b(x^{(-\widetilde a)})-\varepsilon_b(x)\right],
\]
where $\widetilde a$ is an expression- and detection-matched random set. Positive values indicate that programme $a$ contains context associated with prediction of programme $b$ beyond the matched-random expectation. The matrix is directional because ablating $a$ while predicting $b$ differs from ablating $b$ while predicting $a$. It contains 56 ordered, non-self programme pairs. Each pair was evaluated on 24 cells from the target-matched identity, and the null correction is the mean over three independently sampled matched random sets. Figure~6c presents this matrix as a descriptive analysis of directed programme dependence.

\paragraph{Compositional perturbation geometry.}
For perturbation $p$, let $\mathcal I_p$ be its cells and $\mathcal I_0$ the controls. We defined the mean latent displacement
\[
\delta_p=
\frac{1}{|\mathcal I_p|}\sum_{i\in\mathcal I_p}z_i
-
\frac{1}{|\mathcal I_0|}\sum_{i\in\mathcal I_0}z_i.
\]
For a measured double perturbation $(a,b)$, additive consistency was
\[
C_{ab}=
\cos\!\left(\delta_{ab},\delta_a+\delta_b\right).
\]
The null distribution replaced $(a,b)$ with 200 expression-matched random single-perturbation pairs for each combination condition. Each displacement used 24 cells from the combination or single condition and the same 128 control cells. The statistical unit was one of the 80 evaluable combination conditions, and each observed cosine was paired with that condition's random-pair mean. All 80 differences were positive (two-sided paired Wilcoxon signed-rank test, $n=80$ conditions, nominal $P=7.8495\times10^{-15}$). This was the pre-specified aggregate hypothesis for Fig.~6d. Literature-supported pairs highlighted in Fig.~6 include MAP2K3--MAP2K6, CEBPA--CEBPE and TBX2--TBX3\cite{raingeaud1996mkk,avellino2022cebpa,singh2012tbx}; these examples provide descriptive biological context.

\paragraph{Null distributions and multiplicity.}
Expression- and detection-matched nulls were constructed by replacing each candidate gene with genes from the same bins of mean expression and detection frequency. Figure~6b uses one matched set for each of eight programmes; Fig.~6c uses three sets for each source programme as a descriptive matrix; and Fig.~6d uses 200 random single-perturbation pairs for each of 80 combination conditions. Inferential testing comprised the two pre-specified aggregate hypotheses in Fig.~6b,d, which are reported with their nominal two-sided $P$ values.

\begin{algorithm}[H]
\caption{Construction of the biological-knowledge analyses in Fig.~6}
\label{alg:biology}
\begin{algorithmic}[1]
\Require Frozen encoder $f$; hPancreas cells and labels; Norman controls, singles and doubles; graph modules $\mathcal M$
\State Extract token states and cell embeddings once with $f$
\For{each hPancreas cell type $c$}
  \State Count the max-token winner in each held-out cell
  \State Compute winner fractions $A_{cg}$
\EndFor
\For{each programme $B\in\mathcal M$ and cell type $c$}
  \State Re-embed cells after removing $B$
  \State Re-embed cells after removing matched random gene sets
  \State Compute excess displacement $E_{cB}$
\EndFor
\For{each ordered programme pair $(a,b)$}
  \State Ablate context programme $a$ and measure target error for $b$
  \State Subtract the expression-matched random-set effect to obtain $I_{a\rightarrow b}$
\EndFor
\For{each Norman perturbation $p$}
  \State Compute mean displacement $\delta_p$ from controls
\EndFor
\For{each measured double perturbation $(a,b)$}
  \State Compute $C_{ab}=\cos(\delta_{ab},\delta_a+\delta_b)$
  \State Compare with expression-matched random perturbation pairs
\EndFor
\State Apply the two pre-specified aggregate paired tests; retain the directed matrix as descriptive
\State Export one source-data table per panel and render Fig.~6
\end{algorithmic}
\end{algorithm}

\subsection*{Figure source data}

Figures~1--6 were generated from panel-level source-data tables. The conceptual method schematic contains no numerical result and summarizes graph blocks, context removal, the EMA teacher, the stop-gradient block target and aggregate block loss. Figure~6 combines the max-token, programme-ablation, directed-dependency and perturbation-composition analyses described above.

Each quantitative panel was assembled from the stated transformation of its source table. Model colours, marker shapes, dataset order and axis limits were specified consistently across panels. Error bars denote either seed standard deviations or dataset-bootstrap intervals as identified in the captions. The source-data archive contains the numerical input for every panel.

\clearpage

\section*{Supplementary benchmark tables}

The tables below report the complete CellBench-style outputs used in this study. Annotation is a within-dataset few-shot benchmark. \method{} rows use the frozen-embedding and probe protocols defined above. Tables are numbered consecutively as Table S1--S13.

\begin{table*}[htbp]
  \centering
  \suppTableStyle
  \caption{Cross-dataset summary of CellBench within-dataset Top-5 few-shot annotation. Macro-$F_1$ values are averaged over the nine datasets in Tables~\ref{tab:full_annotation_transfer_jepa_noprec_1}--\ref{tab:full_annotation_transfer_jepa_noprec_3}; intervals are percentile 95\% confidence intervals from 20,000 nonparametric bootstrap resamples of datasets. Mean rank is calculated within each dataset among the ten displayed methods. Datasets, rather than cells, are the uncertainty unit.}
  \label{tab:annotation_cross_dataset_summary}
  \begin{tabular}{lrrrr}
    \toprule
    Model & Datasets & Mean macro-$F_1$ (\%) & Dataset-bootstrap 95\% CI & Mean rank \\
    \midrule
    UMAP & 9 & 52.43 & 39.87--65.03 & 6.56 \\
    scVI & 9 & 49.29 & 41.17--56.95 & 6.67 \\
    CellPLM & 9 & 65.27 & 55.21--74.61 & 3.44 \\
    Geneformer & 9 & 40.00 & 32.90--47.34 & 8.33 \\
    LangCell & 9 & 32.26 & 24.35--42.52 & 9.44 \\
    scGPT & 9 & 52.04 & 43.46--61.14 & 7.22 \\
    scMulan & 9 & 63.77 & 54.69--72.95 & 4.00 \\
    scFoundation & 9 & 66.17 & 56.34--75.97 & 3.00 \\
    Nicheformer & 9 & 67.30 & 57.46--77.43 & 2.00 \\
    \biomrow\textbf{GB-JEPA} & 9 & 61.62 & 51.56--71.94 & 4.33 \\
    \bottomrule
  \end{tabular}
\end{table*}

\clearpage
\begin{table*}[htbp]
  \centering
  \suppWideTableStyle
  \caption{Full CellBench within-dataset few-shot annotation benchmark with GB-JEPA. The reported metrics are Acc/F$_1$/Recall (\%).}
  \label{tab:full_annotation_transfer_jepa_noprec_1}
  \resizebox{\textwidth}{!}{%
  \begin{tabular}{llccccccccc}
    \toprule
    Dataset & Model & \multicolumn{3}{c}{Top1} & \multicolumn{3}{c}{Top5} & \multicolumn{3}{c}{Top9} \\
    \cmidrule(lr){3-5}\cmidrule(lr){6-8}\cmidrule(lr){9-11}
      & & Acc & F$_1$ & Recall & Acc & F$_1$ & Recall & Acc & F$_1$ & Recall \\
    \midrule
    PBMC12k & UMAP & 44.73 & 32.17 & 37.52 & 55.20 & 42.51 & 44.60 & 56.51 & 45.94 & 47.38 \\
     & scVI & 23.35 & 21.02 & 25.20 & 50.00 & 50.30 & 47.34 & 62.32 & 60.08 & 56.19 \\
     & CellPLM & 81.79 & 71.95 & 73.04 & 90.44 & 83.02 & 81.57 & 93.04 & 87.19 & 84.77 \\
     & Geneformer & 57.66 & 40.89 & 44.81 & 75.66 & 59.55 & 57.52 & 76.21 & 62.38 & 59.69 \\
     & LangCell & 19.62 & 17.18 & 21.61 & 36.90 & 35.02 & 36.13 & 43.56 & 41.83 & 41.62 \\
     & scGPT & 46.31 & 31.47 & 36.93 & 70.10 & 54.98 & 54.10 & 72.21 & 59.20 & 56.91 \\
     & scMulan & 80.96 & 69.30 & 72.24 & 91.48 & 83.77 & 81.92 & 92.25 & 87.12 & 84.98 \\
     & scFoundation & 78.33 & 67.03 & 67.24 & 91.18 & 85.50 & 83.16 & 91.23 & 87.29 & 84.45 \\
     & Nicheformer & 82.36 & 71.33 & 71.86 & 91.56 & 84.72 & 84.05 & 92.77 & 88.65 & 87.36 \\
    \biomrow & \textbf{GB-JEPA} & 58.11 & 47.47 & 54.52 & 83.31 & 70.95 & 77.94 & 84.12 & 73.95 & 81.45 \\
    \rankrow & \emph{GB-JEPA rank} & 5/10 & 5/10 & 5/10 & 5/10 & 5/10 & 5/10 & 5/10 & 5/10 & 5/10 \\
    \midrule
    hPancreas & UMAP & 65.11 & 55.70 & 59.49 & 87.83 & 65.22 & 65.97 & 83.80 & 64.33 & 65.62 \\
     & scVI & 39.62 & 34.10 & 34.49 & 82.91 & 68.18 & 62.20 & 89.87 & 69.84 & 64.38 \\
     & CellPLM & 57.60 & 55.09 & 56.41 & 75.54 & 65.28 & 62.60 & 83.70 & 68.50 & 65.28 \\
     & Geneformer & 22.31 & 23.65 & 26.97 & 45.92 & 44.46 & 44.15 & 55.75 & 48.96 & 46.44 \\
     & LangCell & 19.21 & 17.13 & 20.24 & 37.63 & 33.84 & 35.67 & 47.74 & 39.85 & 39.30 \\
     & scGPT & 40.17 & 41.60 & 43.25 & 59.93 & 56.01 & 55.12 & 71.79 & 56.17 & 51.71 \\
     & scMulan & 58.21 & 49.20 & 52.00 & 86.82 & 69.72 & 65.98 & 91.20 & 74.98 & 70.66 \\
     & scFoundation & 47.54 & 47.30 & 49.50 & 87.46 & 75.30 & 71.87 & 93.62 & 76.47 & 72.88 \\
     & Nicheformer & 45.95 & 49.92 & 52.84 & 73.94 & 67.96 & 65.74 & 84.28 & 74.36 & 70.74 \\
    \biomrow & \textbf{GB-JEPA} & 58.14 & 43.13 & 64.87 & 93.27 & 75.44 & 88.61 & 96.37 & 80.68 & 90.22 \\
    \rankrow & \emph{GB-JEPA rank} & 3/10 & 6/10 & 1/10 & 1/10 & 1/10 & 1/10 & 1/10 & 1/10 & 1/10 \\
    \midrule
    Cortex & UMAP & 80.38 & 73.06 & 76.98 & 83.50 & 75.10 & 77.84 & 88.34 & 79.06 & 81.70 \\
     & scVI & 40.35 & 35.53 & 34.68 & 77.87 & 61.31 & 55.84 & 93.77 & 79.00 & 73.43 \\
     & CellPLM & 90.52 & 77.65 & 76.22 & 95.76 & 84.48 & 83.13 & 96.91 & 81.69 & 80.31 \\
     & Geneformer & 43.31 & 29.27 & 33.25 & 77.30 & 54.20 & 52.27 & 87.56 & 63.60 & 60.28 \\
     & LangCell & 65.13 & 39.86 & 41.69 & 88.33 & 67.61 & 64.93 & 92.52 & 72.60 & 69.65 \\
     & scGPT & 84.31 & 67.34 & 67.06 & 96.60 & 80.20 & 79.50 & 98.21 & 84.16 & 82.89 \\
     & scMulan & 67.84 & 60.67 & 61.80 & 97.47 & 84.70 & 83.31 & 98.60 & 87.00 & 85.63 \\
     & scFoundation & 98.03 & 85.27 & 84.50 & 97.84 & 88.75 & 88.51 & 98.87 & 92.04 & 92.20 \\
     & Nicheformer & 97.80 & 84.58 & 85.65 & 98.46 & 92.51 & 91.54 & 98.91 & 90.74 & 90.95 \\
    \biomrow & \textbf{GB-JEPA} & 88.39 & 72.57 & 80.81 & 99.07 & 90.34 & 90.39 & 99.09 & 91.06 & 90.93 \\
    \rankrow & \emph{GB-JEPA rank} & 4/10 & 5/10 & 3/10 & 1/10 & 2/10 & 2/10 & 1/10 & 2/10 & 3/10 \\
    \bottomrule
  \end{tabular}}
\end{table*}

\begin{table*}[htbp]
  \centering
  \suppWideTableStyle
  \caption{Full CellBench within-dataset few-shot annotation benchmark with GB-JEPA (continued). The reported metrics are Acc/F$_1$/Recall (\%).}
  \label{tab:full_annotation_transfer_jepa_noprec_2}
  \resizebox{\textwidth}{!}{%
  \begin{tabular}{llccccccccc}
    \toprule
    Dataset & Model & \multicolumn{3}{c}{Top1} & \multicolumn{3}{c}{Top5} & \multicolumn{3}{c}{Top9} \\
    \cmidrule(lr){3-5}\cmidrule(lr){6-8}\cmidrule(lr){9-11}
      & & Acc & F$_1$ & Recall & Acc & F$_1$ & Recall & Acc & F$_1$ & Recall \\
    \midrule
    MS & UMAP & 23.87 & 23.75 & 27.89 & 29.01 & 30.31 & 33.26 & 32.41 & 33.39 & 34.80 \\
     & scVI & 30.91 & 29.55 & 30.94 & 63.60 & 58.70 & 57.29 & 68.69 & 63.69 & 61.79 \\
     & CellPLM & 52.78 & 50.97 & 53.51 & 65.31 & 65.25 & 65.09 & 69.28 & 68.02 & 68.12 \\
     & Geneformer & 23.57 & 26.32 & 29.19 & 32.85 & 37.13 & 37.21 & 35.05 & 38.98 & 38.82 \\
     & LangCell & 13.02 & 10.37 & 12.60 & 18.28 & 16.47 & 17.54 & 20.67 & 19.15 & 20.03 \\
     & scGPT & 32.93 & 29.94 & 33.89 & 52.09 & 49.14 & 49.28 & 59.30 & 56.38 & 56.51 \\
     & scMulan & 37.75 & 35.40 & 37.81 & 55.56 & 55.92 & 56.40 & 59.09 & 59.65 & 60.10 \\
     & scFoundation & 41.77 & 42.94 & 45.54 & 53.36 & 56.62 & 56.56 & 58.55 & 61.12 & 61.61 \\
     & Nicheformer & 43.42 & 42.70 & 45.97 & 60.13 & 58.02 & 58.37 & 63.71 & 62.06 & 63.45 \\
    \biomrow & \textbf{GB-JEPA} & 33.69 & 31.84 & 34.37 & 51.82 & 51.84 & 56.82 & 58.01 & 57.33 & 61.64 \\
    \rankrow & \emph{GB-JEPA rank} & 5/10 & 5/10 & 5/10 & 7/10 & 6/10 & 4/10 & 7/10 & 6/10 & 4/10 \\
    \midrule
    Liver & UMAP & 62.61 & 62.15 & 65.56 & 70.36 & 75.49 & 74.48 & 70.33 & 74.77 & 73.06 \\
     & scVI & 25.43 & 23.02 & 24.57 & 41.86 & 43.41 & 40.96 & 52.87 & 52.71 & 49.55 \\
     & CellPLM & 56.93 & 56.74 & 56.04 & 70.43 & 72.34 & 69.71 & 74.24 & 75.04 & 72.05 \\
     & Geneformer & 26.32 & 22.58 & 26.04 & 41.91 & 37.88 & 38.36 & 51.81 & 47.88 & 46.42 \\
     & LangCell & 22.30 & 17.57 & 22.48 & 34.36 & 30.34 & 32.57 & 43.89 & 39.26 & 40.40 \\
     & scGPT & 41.44 & 36.27 & 40.06 & 58.96 & 56.19 & 55.47 & 66.08 & 62.26 & 59.57 \\
     & scMulan & 55.31 & 52.15 & 53.54 & 65.93 & 65.59 & 62.16 & 69.70 & 69.58 & 66.66 \\
     & scFoundation & 53.87 & 50.99 & 50.91 & 70.49 & 68.45 & 65.64 & 72.67 & 72.62 & 69.64 \\
     & Nicheformer & 49.39 & 46.53 & 47.88 & 72.89 & 73.91 & 71.52 & 74.06 & 75.68 & 73.25 \\
    \biomrow & \textbf{GB-JEPA} & 35.31 & 31.04 & 41.40 & 63.49 & 63.33 & 71.64 & 70.12 & 69.65 & 77.67 \\
    \rankrow & \emph{GB-JEPA rank} & 7/10 & 7/10 & 6/10 & 6/10 & 6/10 & 2/10 & 5/10 & 5/10 & 1/10 \\
    \midrule
    Zheng68k & UMAP & 23.92 & 22.01 & 25.12 & 31.34 & 28.56 & 30.67 & 32.79 & 29.76 & 31.46 \\
     & scVI & 13.82 & 13.97 & 16.68 & 27.01 & 25.06 & 25.95 & 34.79 & 32.17 & 32.61 \\
     & CellPLM & 44.68 & 36.59 & 40.10 & 47.21 & 43.13 & 44.55 & 48.42 & 46.31 & 48.00 \\
     & Geneformer & 29.18 & 27.31 & 31.76 & 38.41 & 36.97 & 37.28 & 39.77 & 38.44 & 38.78 \\
     & LangCell & 20.57 & 19.97 & 24.14 & 35.63 & 33.44 & 33.57 & 38.03 & 36.95 & 37.19 \\
     & scGPT & 25.24 & 25.00 & 28.26 & 35.47 & 32.81 & 33.15 & 39.40 & 37.49 & 37.90 \\
     & scMulan & 28.80 & 27.78 & 32.49 & 46.69 & 42.63 & 43.87 & 50.31 & 46.80 & 48.06 \\
     & scFoundation & 37.00 & 32.52 & 35.76 & 44.25 & 42.88 & 43.57 & 48.53 & 46.98 & 47.89 \\
     & Nicheformer & 37.98 & 35.17 & 38.52 & 44.64 & 43.26 & 44.66 & 46.99 & 45.66 & 46.37 \\
    \biomrow & \textbf{GB-JEPA} & 26.93 & 23.79 & 30.69 & 40.89 & 38.09 & 44.05 & 41.04 & 40.29 & 47.10 \\
    \rankrow & \emph{GB-JEPA rank} & 6/10 & 7/10 & 6/10 & 5/10 & 5/10 & 3/10 & 5/10 & 5/10 & 4/10 \\
    \bottomrule
  \end{tabular}}
\end{table*}

\begin{table*}[htbp]
  \centering
  \suppWideTableStyle
  \caption{Full CellBench within-dataset few-shot annotation benchmark with GB-JEPA (continued). The reported metrics are Acc/F$_1$/Recall (\%).}
  \label{tab:full_annotation_transfer_jepa_noprec_3}
  \resizebox{\textwidth}{!}{%
  \begin{tabular}{llccccccccc}
    \toprule
    Dataset & Model & \multicolumn{3}{c}{Top1} & \multicolumn{3}{c}{Top5} & \multicolumn{3}{c}{Top9} \\
    \cmidrule(lr){3-5}\cmidrule(lr){6-8}\cmidrule(lr){9-11}
      & & Acc & F$_1$ & Recall & Acc & F$_1$ & Recall & Acc & F$_1$ & Recall \\
    \midrule
    Lung & UMAP & 74.50 & 58.40 & 62.42 & 89.98 & 76.39 & 73.48 & 88.23 & 73.80 & 71.66 \\
     & scVI & 25.90 & 19.21 & 22.93 & 59.59 & 45.57 & 42.98 & 72.14 & 56.42 & 52.33 \\
     & CellPLM & 77.11 & 54.85 & 58.07 & 91.39 & 74.96 & 72.43 & 91.55 & 76.04 & 72.91 \\
     & Geneformer & 24.76 & 17.27 & 23.30 & 53.99 & 40.45 & 41.41 & 63.96 & 50.23 & 49.48 \\
     & LangCell & 23.38 & 15.40 & 20.32 & 41.56 & 31.23 & 34.22 & 45.25 & 34.41 & 36.84 \\
     & scGPT & 49.40 & 33.66 & 37.31 & 80.11 & 59.70 & 57.28 & 85.46 & 67.44 & 63.62 \\
     & scMulan & 61.02 & 46.18 & 48.33 & 84.98 & 68.24 & 64.16 & 88.05 & 72.29 & 67.62 \\
     & scFoundation & 66.31 & 51.45 & 52.30 & 88.22 & 72.43 & 69.77 & 90.74 & 75.42 & 72.35 \\
     & Nicheformer & 59.49 & 50.03 & 55.72 & 91.87 & 77.26 & 74.38 & 92.70 & 78.36 & 75.16 \\
    \biomrow & \textbf{GB-JEPA} & 48.91 & 35.20 & 54.73 & 85.96 & 69.44 & 88.20 & 88.67 & 74.01 & 90.84 \\
    \rankrow & \emph{GB-JEPA rank} & 7/10 & 6/10 & 4/10 & 5/10 & 5/10 & 1/10 & 4/10 & 4/10 & 1/10 \\
    \midrule
    Covid & UMAP & 29.83 & 31.52 & 37.54 & 35.63 & 33.79 & 37.01 & 47.57 & 45.98 & 45.69 \\
     & scVI & 23.27 & 18.49 & 18.50 & 46.21 & 40.34 & 36.23 & 50.86 & 42.21 & 37.43 \\
     & CellPLM & 28.16 & 26.95 & 28.91 & 38.89 & 39.38 & 39.51 & 44.55 & 44.14 & 42.97 \\
     & Geneformer & 13.11 & 10.38 & 12.31 & 24.68 & 22.40 & 22.48 & 27.18 & 24.26 & 24.10 \\
     & LangCell & 10.20 & 7.96 & 10.18 & 23.56 & 17.55 & 18.48 & 29.45 & 22.58 & 21.70 \\
     & scGPT & 20.90 & 17.20 & 19.53 & 37.09 & 32.22 & 33.31 & 45.43 & 38.27 & 36.29 \\
     & scMulan & 33.36 & 28.43 & 28.46 & 50.81 & 45.18 & 42.45 & 60.71 & 50.18 & 46.30 \\
     & scFoundation & 29.81 & 27.16 & 28.10 & 51.62 & 47.90 & 45.89 & 59.79 & 52.19 & 48.63 \\
     & Nicheformer & 32.77 & 30.20 & 30.85 & 54.47 & 49.39 & 47.53 & 57.43 & 50.23 & 46.75 \\
    \biomrow & \textbf{GB-JEPA} & 27.58 & 24.64 & 40.83 & 48.16 & 43.97 & 64.04 & 52.77 & 46.58 & 67.54 \\
    \rankrow & \emph{GB-JEPA rank} & 6/10 & 6/10 & 1/10 & 4/10 & 4/10 & 1/10 & 4/10 & 4/10 & 1/10 \\
    \midrule
    Immune & UMAP & 28.09 & 31.14 & 38.08 & 44.84 & 44.47 & 46.86 & 51.92 & 48.02 & 49.20 \\
     & scVI & 27.16 & 24.79 & 25.26 & 55.21 & 50.77 & 47.91 & 63.23 & 61.10 & 58.73 \\
     & CellPLM & 45.90 & 42.89 & 43.12 & 60.17 & 59.59 & 57.53 & 63.78 & 63.76 & 61.32 \\
     & Geneformer & 18.63 & 13.85 & 15.84 & 31.22 & 26.96 & 26.14 & 33.38 & 32.67 & 32.72 \\
     & LangCell & 14.91 & 12.71 & 15.09 & 26.31 & 24.87 & 25.09 & 30.46 & 30.62 & 30.76 \\
     & scGPT & 29.27 & 25.41 & 26.70 & 50.04 & 47.09 & 44.79 & 55.25 & 53.06 & 50.73 \\
     & scMulan & 35.34 & 36.11 & 37.24 & 59.94 & 58.18 & 55.98 & 63.40 & 63.40 & 62.15 \\
     & scFoundation & 35.68 & 33.92 & 35.37 & 58.72 & 57.68 & 56.93 & 66.52 & 65.46 & 63.41 \\
     & Nicheformer & 40.79 & 37.97 & 38.82 & 61.28 & 58.67 & 57.30 & 65.61 & 64.84 & 62.91 \\
    \biomrow & \textbf{GB-JEPA} & 31.59 & 28.24 & 38.93 & 54.11 & 51.21 & 62.37 & 57.61 & 58.24 & 66.70 \\
    \rankrow & \emph{GB-JEPA rank} & 5/10 & 6/10 & 2/10 & 6/10 & 5/10 & 1/10 & 6/10 & 6/10 & 1/10 \\
    \bottomrule
  \end{tabular}}
\end{table*}

\clearpage
\begin{table*}[htbp]
  \centering
  \suppWideTableStyle
  \caption{CellBench clustering performance with GB-JEPA (AvgBio and accuracy, \%). AvgBio is the mean of NMI, ARI and ASW.}
  \label{tab:clustering_performance_jepa}
  \resizebox{\textwidth}{!}{%
  \begin{tabular}{lcccccccccc}
    \toprule
    Model
      & \multicolumn{2}{c}{PBMC12k} & \multicolumn{2}{c}{hPancreas} & \multicolumn{2}{c}{Cortex} & \multicolumn{2}{c}{MS} & \multicolumn{2}{c}{Liver} \\
    \cmidrule(lr){2-3}\cmidrule(lr){4-5}\cmidrule(lr){6-7}\cmidrule(lr){8-9}\cmidrule(lr){10-11}
      & AvgBio & Acc & AvgBio & Acc & AvgBio & Acc & AvgBio & Acc & AvgBio & Acc \\
    \midrule
    UMAP & 58.59 & 70.51 & \textbf{65.86} & 56.57 & 85.35 & 94.28 & 33.31 & 36.53 & 52.70 & 48.11 \\
    scVI & 71.68 & 89.12 & 57.27 & 60.63 & 84.62 & 98.75 & \textbf{58.56} & \textbf{64.76} & 59.62 & 70.92 \\
    CellPLM & \textbf{80.62} & \textbf{92.51} & 64.17 & \textbf{76.16} & \textbf{87.24} & 97.02 & 52.94 & 55.90 & \textbf{67.99} & \textbf{80.98} \\
    Geneformer & 59.50 & 74.75 & 33.78 & 43.22 & 72.71 & 93.77 & 32.23 & 34.38 & 28.08 & 30.17 \\
    LangCell & 26.77 & 18.22 & 31.64 & 37.64 & 77.33 & 94.30 & 24.10 & 23.39 & 26.70 & 21.01 \\
    scGPT & 55.59 & 71.77 & 45.29 & 52.58 & 84.56 & 96.64 & 42.47 & 38.37 & 61.18 & 79.38 \\
    scMulan & 69.55 & 80.18 & 56.81 & 58.90 & 68.87 & 74.56 & 43.00 & 49.40 & 38.86 & 33.60 \\
    scFoundation & 71.81 & 84.11 & 47.95 & 45.04 & 77.44 & 85.77 & 46.41 & 47.21 & 61.46 & 76.48 \\
    \biomrow\textbf{GB-JEPA} & 71.39 & 89.43 & 61.79 & 67.93 & 85.67 & \textbf{99.10} & 42.76 & 45.19 & 58.23 & 68.52 \\
    \rankrow\emph{GB-JEPA rank} & 4/9 & 2/9 & 3/9 & 2/9 & 2/9 & 1/9 & 5/9 & 5/9 & 5/9 & 5/9 \\
    \bottomrule
  \end{tabular}
  }

  \vspace{0.3cm}

  \resizebox{\textwidth}{!}{%
  \begin{tabular}{lcccccccc}
    \toprule
    Model
      & \multicolumn{2}{c}{Zheng68k} & \multicolumn{2}{c}{Lung} & \multicolumn{2}{c}{Covid} & \multicolumn{2}{c}{Immune} \\
    \cmidrule(lr){2-3}\cmidrule(lr){4-5}\cmidrule(lr){6-7}\cmidrule(lr){8-9}
      & AvgBio & Acc & AvgBio & Acc & AvgBio & Acc & AvgBio & Acc \\
    \midrule
    UMAP & 32.73 & 38.01 & 65.44 & 61.52 & 40.30 & 38.28 & 44.76 & 41.51 \\
    scVI & 41.16 & 55.31 & 71.28 & 89.11 & 43.83 & 40.01 & 51.79 & 46.76 \\
    CellPLM & \textbf{42.60} & \textbf{57.16} & \textbf{79.12} & \textbf{92.71} & \textbf{48.62} & 48.52 & \textbf{58.63} & \textbf{63.13} \\
    Geneformer & 39.17 & 47.14 & 36.97 & 44.98 & 34.05 & 34.19 & 41.03 & 39.71 \\
    LangCell & 33.08 & 36.92 & 34.21 & 55.38 & 26.30 & 21.93 & 35.22 & 28.33 \\
    scGPT & 37.68 & 43.83 & 70.98 & 88.25 & 46.46 & \textbf{51.46} & 50.59 & 49.01 \\
    scMulan & 36.16 & 42.80 & 53.27 & 56.87 & 44.53 & 39.59 & 45.88 & 36.25 \\
    scFoundation & 39.59 & 51.54 & 70.35 & 80.58 & 46.45 & 46.28 & 47.71 & 42.06 \\
    \biomrow\textbf{GB-JEPA} & 40.96 & 47.93 & 71.95 & 89.48 & 45.75 & 47.99 & 50.61 & 50.69 \\
    \rankrow\emph{GB-JEPA rank} & 3/9 & 4/9 & 2/9 & 2/9 & 4/9 & 3/9 & 3/9 & 2/9 \\
    \bottomrule
  \end{tabular}
  }
\end{table*}

\clearpage
\begin{table*}[htbp]
  \centering
  \suppDenseTableStyle
  \caption{CellBench perturbation-prediction error metrics with GB-JEPA. Values are mean$\pm$s.d. over five seeds; GB-JEPA ranks are computed among the nine displayed models.}
  \label{tab:perturbation_jepa_error_rank}
  \begin{tabular}{llrrrrrrrr}
    \toprule
    Dataset & Model & \multicolumn{4}{c}{MSE-LFC $\downarrow$} & \multicolumn{4}{c}{MSE-Expr $\downarrow$} \\
    \cmidrule(lr){3-6} \cmidrule(lr){7-10}
     &  & Top1 & Top5 & Top9 & Mean & Top1 & Top5 & Top9 & Mean \\
    \midrule
    Adamson & UMAP & 15.446$\pm$0.140 & 15.257$\pm$0.049 & 15.067$\pm$0.052 & 15.234 & 0.218$\pm$0.003 & 0.201$\pm$0.001 & 0.196$\pm$0.001 & 0.203 \\
     & scVI & 15.405$\pm$0.081 & 15.378$\pm$0.071 & 15.173$\pm$0.051 & 15.303 & 0.217$\pm$0.001 & 0.197$\pm$0.001 & 0.194$\pm$0.001 & 0.202 \\
     & CellPLM & 15.324$\pm$0.033 & 15.346$\pm$0.037 & 15.165$\pm$0.079 & 15.272 & 0.225$\pm$0.002 & 0.201$\pm$0.000 & 0.196$\pm$0.000 & 0.206 \\
     & Geneformer & 15.333$\pm$0.040 & 15.432$\pm$0.043 & 15.316$\pm$0.048 & 15.374 & 0.224$\pm$0.002 & 0.204$\pm$0.001 & 0.200$\pm$0.001 & 0.208 \\
     & LangCell & 15.339$\pm$0.032 & 15.405$\pm$0.056 & 15.248$\pm$0.050 & 15.334 & 0.225$\pm$0.002 & 0.204$\pm$0.001 & 0.199$\pm$0.001 & 0.208 \\
     & scGPT & 15.348$\pm$0.087 & 15.396$\pm$0.018 & 15.241$\pm$0.086 & 15.337 & 0.224$\pm$0.002 & 0.203$\pm$0.001 & 0.199$\pm$0.001 & 0.207 \\
     & scMulan & 15.431$\pm$0.244 & 15.286$\pm$0.068 & 15.286$\pm$0.047 & 15.268 & 0.207$\pm$0.002 & 0.199$\pm$0.001 & 0.196$\pm$0.001 & 0.200 \\
     & scFoundation & 15.928$\pm$0.300 & 15.224$\pm$0.123 & 15.193$\pm$0.084 & 15.388 & 0.202$\pm$0.001 & 0.195$\pm$0.001 & 0.194$\pm$0.001 & 0.196 \\
    \biomrow & \textbf{GB-JEPA} & \textbf{14.887$\pm$0.171} & \textbf{14.381$\pm$0.076} & \textbf{14.801$\pm$0.080} & \textbf{14.694} & \textbf{0.198$\pm$0.001} & \textbf{0.191$\pm$0.000} & \textbf{0.192$\pm$0.000} & \textbf{0.194} \\
    \rankrow & \emph{GB-JEPA rank} & 1/9 & 1/9 & 1/9 & 1/9 & 1/9 & 1/9 & 1/9 & 1/9 \\
    \midrule
    Norman & UMAP & 6.528$\pm$0.048 & 6.535$\pm$0.082 & 6.473$\pm$0.036 & 6.507 & 0.089$\pm$0.000 & 0.085$\pm$0.001 & 0.084$\pm$0.000 & 0.085 \\
     & scVI & 6.633$\pm$0.062 & 6.406$\pm$0.043 & 6.457$\pm$0.116 & 6.423 & 0.088$\pm$0.001 & 0.083$\pm$0.000 & 0.083$\pm$0.000 & 0.084 \\
     & CellPLM & 6.462$\pm$0.026 & 6.501$\pm$0.042 & 6.460$\pm$0.058 & 6.477 & 0.090$\pm$0.000 & 0.084$\pm$0.000 & 0.083$\pm$0.000 & 0.085 \\
     & Geneformer & 6.467$\pm$0.037 & 6.535$\pm$0.054 & 6.465$\pm$0.021 & 6.498 & 0.090$\pm$0.000 & 0.085$\pm$0.001 & 0.083$\pm$0.000 & 0.086 \\
     & LangCell & 6.484$\pm$0.016 & 6.531$\pm$0.033 & 6.435$\pm$0.022 & 6.499 & 0.090$\pm$0.001 & 0.085$\pm$0.000 & 0.083$\pm$0.000 & 0.086 \\
     & scGPT & 6.457$\pm$0.014 & 6.541$\pm$0.040 & 6.480$\pm$0.021 & 6.497 & 0.089$\pm$0.001 & 0.085$\pm$0.001 & 0.084$\pm$0.000 & 0.085 \\
     & scMulan & 6.451$\pm$0.074 & 6.378$\pm$0.087 & 6.285$\pm$0.089 & 6.330 & 0.086$\pm$0.001 & 0.083$\pm$0.000 & 0.081$\pm$0.000 & 0.083 \\
     & scFoundation & 6.737$\pm$0.149 & 6.523$\pm$0.062 & 6.343$\pm$0.077 & 6.461 & 0.086$\pm$0.001 & 0.083$\pm$0.000 & 0.081$\pm$0.000 & 0.083 \\
    \biomrow & \textbf{GB-JEPA} & \textbf{6.236$\pm$0.046} & \textbf{6.249$\pm$0.070} & \textbf{6.200$\pm$0.048} & \textbf{6.188} & \textbf{0.083$\pm$0.000} & \textbf{0.082$\pm$0.000} & \textbf{0.081$\pm$0.000} & \textbf{0.082} \\
    \rankrow & \emph{GB-JEPA rank} & 1/9 & 1/9 & 1/9 & 1/9 & 1/9 & 1/9 & 1/9 & 1/9 \\
    \midrule
    \bottomrule
  \end{tabular}
\end{table*}

\clearpage

\begin{table*}[htbp]
  \centering
  \suppDenseTableStyle
  \caption{CellBench perturbation-prediction directional and response-gene metrics with GB-JEPA. Values are mean$\pm$s.d. over five seeds; GB-JEPA ranks are computed among the nine displayed models.}
  \label{tab:perturbation_jepa_agreement_rank}
  \begin{tabular}{llrrrrrrrr}
    \toprule
    Dataset & Model & \multicolumn{4}{c}{Sign $\uparrow$} & \multicolumn{4}{c}{Top50 $\uparrow$} \\
    \cmidrule(lr){3-6} \cmidrule(lr){7-10}
     &  & Top1 & Top5 & Top9 & Mean & Top1 & Top5 & Top9 & Mean \\
    \midrule
    Adamson & UMAP & 0.735$\pm$0.004 & 0.739$\pm$0.001 & 0.740$\pm$0.001 & 0.739 & 0.073$\pm$0.001 & 0.085$\pm$0.003 & 0.092$\pm$0.002 & 0.084 \\
     & scVI & 0.736$\pm$0.002 & 0.738$\pm$0.002 & 0.741$\pm$0.001 & 0.739 & 0.074$\pm$0.003 & 0.094$\pm$0.001 & 0.105$\pm$0.002 & 0.092 \\
     & CellPLM & 0.740$\pm$0.001 & 0.739$\pm$0.001 & 0.739$\pm$0.002 & 0.740 & 0.072$\pm$0.001 & 0.088$\pm$0.002 & 0.097$\pm$0.002 & 0.086 \\
     & Geneformer & 0.741$\pm$0.001 & 0.741$\pm$0.000 & 0.741$\pm$0.002 & 0.741 & 0.072$\pm$0.001 & 0.086$\pm$0.002 & 0.093$\pm$0.003 & 0.083 \\
     & LangCell & 0.740$\pm$0.001 & 0.740$\pm$0.001 & 0.740$\pm$0.001 & 0.740 & 0.073$\pm$0.001 & 0.083$\pm$0.002 & 0.089$\pm$0.001 & 0.081 \\
     & scGPT & 0.740$\pm$0.001 & 0.741$\pm$0.001 & 0.742$\pm$0.002 & 0.741 & 0.074$\pm$0.002 & 0.086$\pm$0.002 & 0.095$\pm$0.001 & 0.085 \\
     & scMulan & 0.737$\pm$0.005 & 0.740$\pm$0.001 & 0.738$\pm$0.001 & 0.740 & 0.077$\pm$0.004 & 0.097$\pm$0.001 & 0.107$\pm$0.002 & 0.096 \\
     & scFoundation & 0.730$\pm$0.006 & 0.741$\pm$0.002 & 0.742$\pm$0.001 & 0.739 & \textbf{0.087$\pm$0.004} & 0.102$\pm$0.002 & 0.109$\pm$0.003 & 0.100 \\
    \biomrow & \textbf{GB-JEPA} & \textbf{0.746$\pm$0.002} & \textbf{0.751$\pm$0.001} & \textbf{0.745$\pm$0.002} & \textbf{0.748} & 0.086$\pm$0.003 & \textbf{0.113$\pm$0.002} & \textbf{0.110$\pm$0.003} & \textbf{0.104} \\
    \rankrow & \emph{GB-JEPA rank} & 1/9 & 1/9 & 1/9 & 1/9 & 2/9 & 1/9 & 1/9 & 1/9 \\
    \midrule
    Norman & UMAP & 0.885$\pm$0.002 & 0.886$\pm$0.001 & 0.887$\pm$0.001 & 0.886 & 0.093$\pm$0.005 & 0.117$\pm$0.005 & 0.125$\pm$0.004 & 0.113 \\
     & scVI & 0.883$\pm$0.001 & 0.889$\pm$0.001 & 0.888$\pm$0.002 & 0.888 & 0.102$\pm$0.002 & 0.130$\pm$0.006 & 0.136$\pm$0.007 & 0.127 \\
     & CellPLM & 0.887$\pm$0.001 & 0.886$\pm$0.001 & 0.887$\pm$0.001 & 0.887 & 0.097$\pm$0.003 & 0.115$\pm$0.004 & 0.126$\pm$0.004 & 0.114 \\
     & Geneformer & 0.888$\pm$0.001 & 0.887$\pm$0.001 & 0.888$\pm$0.000 & 0.887 & 0.098$\pm$0.001 & 0.114$\pm$0.005 & 0.126$\pm$0.002 & 0.114 \\
     & LangCell & 0.888$\pm$0.001 & 0.887$\pm$0.001 & 0.889$\pm$0.000 & 0.887 & 0.097$\pm$0.002 & 0.111$\pm$0.003 & 0.127$\pm$0.001 & 0.113 \\
     & scGPT & 0.888$\pm$0.000 & 0.886$\pm$0.000 & 0.887$\pm$0.001 & 0.887 & 0.099$\pm$0.004 & 0.117$\pm$0.002 & 0.128$\pm$0.005 & 0.116 \\
     & scMulan & 0.888$\pm$0.001 & 0.888$\pm$0.002 & 0.889$\pm$0.002 & 0.889 & 0.115$\pm$0.003 & 0.131$\pm$0.004 & 0.143$\pm$0.004 & 0.132 \\
     & scFoundation & 0.883$\pm$0.003 & 0.887$\pm$0.001 & 0.890$\pm$0.002 & 0.888 & 0.116$\pm$0.005 & 0.128$\pm$0.004 & 0.142$\pm$0.005 & 0.131 \\
    \biomrow & \textbf{GB-JEPA} & \textbf{0.889$\pm$0.002} & \textbf{0.891$\pm$0.001} & \textbf{0.891$\pm$0.001} & \textbf{0.891} & \textbf{0.134$\pm$0.006} & \textbf{0.134$\pm$0.003} & 0.142$\pm$0.001 & \textbf{0.137} \\
    \rankrow & \emph{GB-JEPA rank} & 1/9 & 1/9 & 1/9 & 1/9 & 1/9 & 1/9 & 2/9 & 1/9 \\
    \midrule
    \bottomrule
  \end{tabular}
\end{table*}

\end{document}